%% file: main.tex
\documentclass{article} 

\PassOptionsToPackage{numbers, compress}{natbib}
\usepackage[preprint]{neurips_2022}
\usepackage[utf8]{inputenc} 
\usepackage[T1]{fontenc}    
\usepackage[hidelinks]{hyperref}       
\usepackage{url}            
\usepackage{booktabs}       
\usepackage{amsfonts}       
\usepackage{nicefrac}       
\usepackage{microtype}      
\input{math_commands.tex}

\input{math_commands_cc.tex}
\usepackage{url}
\usepackage[british]{babel}
\usepackage{epstopdf}
\usepackage{csquotes}
\usepackage{amsmath,amssymb,amsthm}
\usepackage{natbib}
\usepackage[dvipsnames]{xcolor}
\usepackage{algorithm}
\usepackage{algorithmic}
\usepackage{wrapfig}
\usepackage{lipsum}

\usepackage{subcaption}
\usepackage{graphicx}
\usepackage{enumitem}

\author{%
  \makebox[.35\linewidth]{Yikai Zhang} \\
  ML Research \\
  Morgan Stanley \\
  \And
  \makebox[.4\linewidth]{Jiachen Yao} \\
  Department of Computer Science,  \\
  Stony Brook University \\
  \AND
  Yusu Wang \\
  Hal{\i}c{\i}o\u{g}lu Data Science Institute,  \\
   University of California, San Diego \\
  \And
  Chao Chen \\
  Department of Biomedical Informatics,  \\
  Stony Brook University \\
}

\newtheorem{thm}{Theorem}
\newtheorem{lemma}{Lemma}

\theoremstyle{definition}
\newtheorem{definition}{Definition}
\newtheorem{remark}{Remark}

\newcommand{\myparagraph}[1]{\noindent\textbf{#1}}

\title{On the Convergence of Optimizing Persistent-Homology-Based Losses }
\begin{document}

	\maketitle
	
	\begin{abstract}
Topological loss based on persistent homology has shown promise in various applications. A topological loss enforces the model to achieve certain desired topological property.
 Despite its empirical success, less is known about the optimization behavior of the loss. In fact, the topological loss involves combinatorial configurations that may  oscillate during optimization. 
In this paper, we introduce a general purpose regularized topology-aware loss. We propose a novel regularization term and also modify existing topological loss. These contributions lead to a new loss function that not only enforces the model to have desired topological behavior, but also achieves satisfying convergence behavior. Our main theoretical result guarantees that the loss can be optimized  efficiently, under mild assumptions. 
    \end{abstract}
	
	\section{Introduction}
Topological data analysis \cite{edelsbrunner2010computational,carlsson2009topology,dey2022computational} characterizes high-order topological information of data via the mathematics of algebraic topology \cite{munkers1984elements,hatcher2000algebraic}. It captures topological structures including connected components, loops, voids, and their high-dimensional analogs in a multi-scale and robust manner. Such topological summary has shown promise in various domains such as image analysis \cite{wu2017optimal,wang2021topotxr}, graph learning \cite{hofer2017deep,hofer2020graph,zhao2019learning,zhao2020persistence,yan2021link}, biomedicine \cite{aukerman2020persistent,lawson2019persistent} and robust machine learning \cite{wu2020topological,zheng2021topological,hu2022trigger}.

Particularly encouraging is the recent invention of the \emph{topology-aware loss functions}, based on the theory of persistent homology \cite{edelsbrunner2000topological,zomorodian2005computing}. These losses are usually written in the form of $L_{supv}+L_{topo}$. The first term is the standard supervision loss, e.g., cross-entropy loss, mean squared error, etc. Whereas the second term is the \emph{topological loss term}. It involves computing and matching persistence diagrams, which summarize the topological information in view of a model's output.
A topological loss enforces the output to have desired topological characteristics, which cannot be articulated by other standard supervision losses. For example, in image analysis, topological losses have been introduced to enforce segmentation models to be correct in topology \cite{hu2019topology,clough2020topological}, and to ensure generative models learn the topology from data \cite{wang2020topogan,gabrielsson2020topology}. In graph learning, topological losses have been introduced to learn task-driven topological representation of the graph \cite{horn2021topological}. Topology-aware losses have also been introduced to regularize classifiers \cite{chen2019topological} and to improve representation learning \cite{hofer2019connectivity,hofer2020topologically}. 

\begin{wrapfigure}{R}{.2\textwidth}
\centering
\includegraphics[height=.12\textheight]{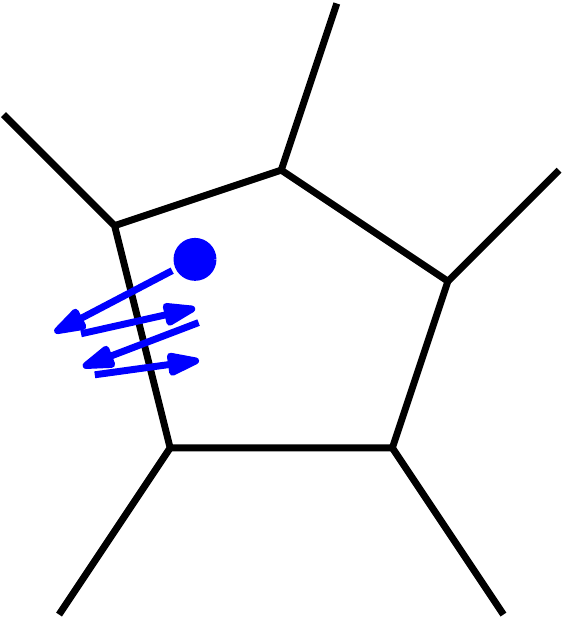}
\caption{An illustration of the optimization of a topological loss.}
\label{fig:optimization}
\end{wrapfigure}
Despite the promising applications of the topological loss and some efforts in accelerating it in practice \cite{solomon2021fast,hu2021topology}, its behavior is far from being understood, especially in terms of optimization. The loss is not guaranteed to decrease after each gradient descent iteration. The topological summary, namely, the persistence diagram, is a highly nonlinear transformation of the data, involving identifying and pairing critical points of certain scalar function generated by the model. The loss function also involves comparing two persistence diagrams via an optimal matching. All these combinatorial information can change unpredictably as we continuously update the model via gradient descent. 
As illustrated in Fig.~\ref{fig:optimization}, the parameter space is partitioned into regions, each of which has a unique combinatorial configuration of the pairing of critical points and the matching of persistence diagrams. A gradient descent step (blue arrow) can unknowingly enter a different configuration, resulting in a higher topological loss.
Carri\'{e}re et al.~\cite{carriere2021optimizing} show that the topological loss can eventually converge. But it is unclear how efficient the optimization can be. It is possible that the loss may take exponential time to converge because of the oscillation of the underlying combinatorial configuration.

In this paper, we provide a theoretical guarantee on the convergence rate of such losses. Instead of studying the topological loss alone, we look at the complete loss function as a whole, which includes both a standard supervision loss and the topological loss. We propose two modifications to the loss function: (1) adding a regularization term based on total persistence, which is essentially the ``norm'' of a persistence diagram; (2) modify the topological loss and drop the part of the loss contributing to removing (noise) topological structures, as this has already been covered by the regularization term/total persistence. 

These two modifications allow us to design a novel general purpose \emph{regularized topology-aware loss function}, which not only preserves the desired topological property of the output, but also achieves a theoretically guaranteed convergence rate. Our main theorem shows that with mild assumptions, with a carefully chosen learning rate, the loss function can be optimized with satisfying convergence rate, $O(1/\varepsilon)$.

	\begin{figure}[b]
\centering
  \begin{tabular}{@{\hskip -0.1in}c@{\hskip -0.05in}c@{\hskip -0.01in}ccc}
    \includegraphics[height=.1\textheight]{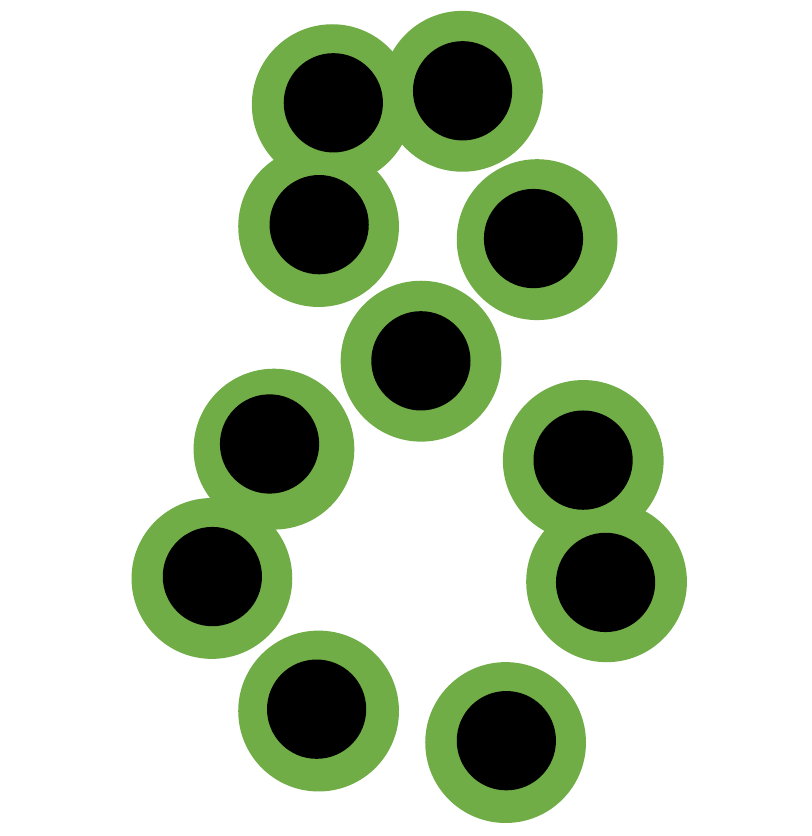} &
    \includegraphics[height=.1\textheight]{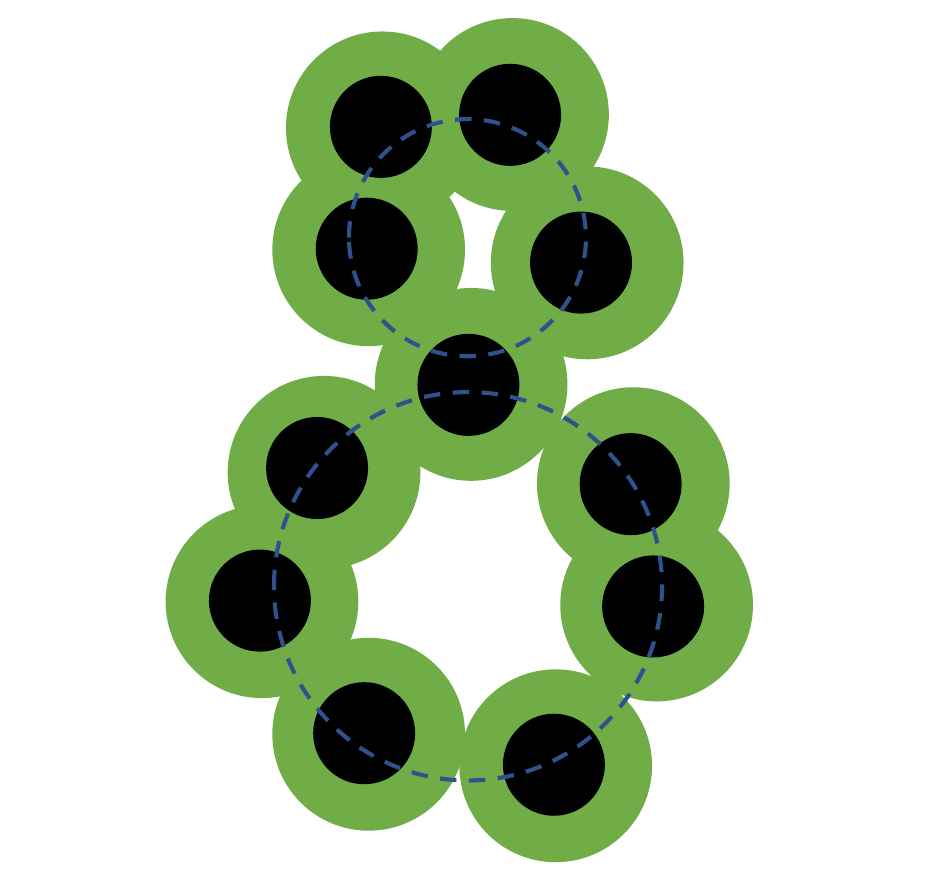} &
    \includegraphics[height=.1\textheight]{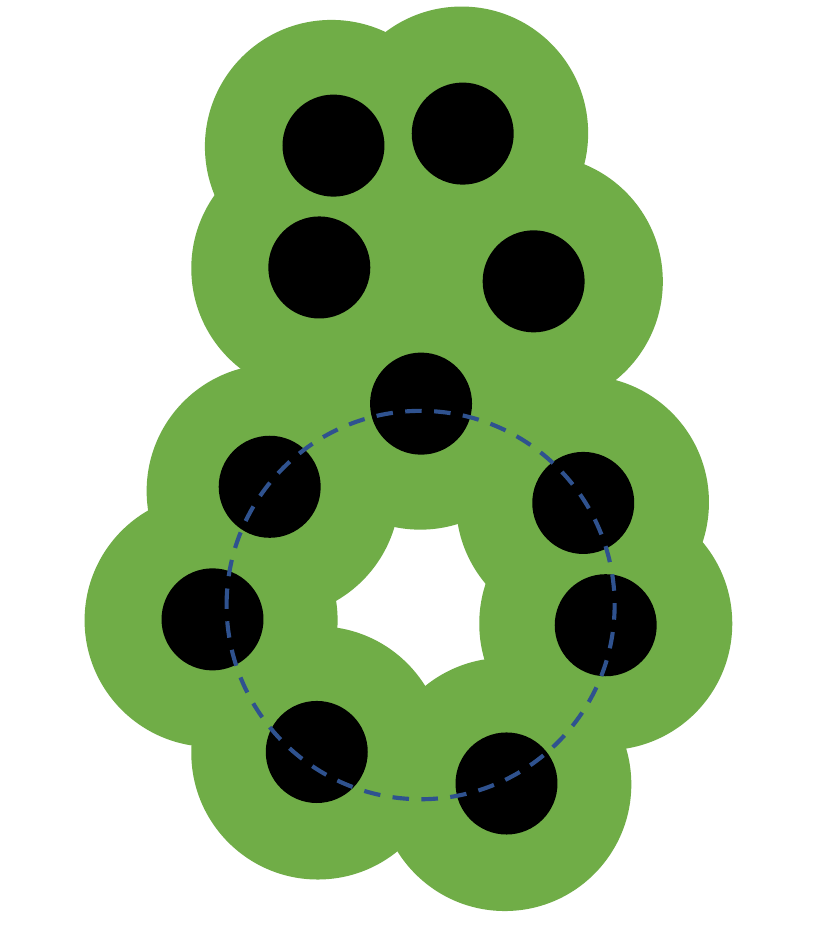} &
    \includegraphics[height=.1\textheight]{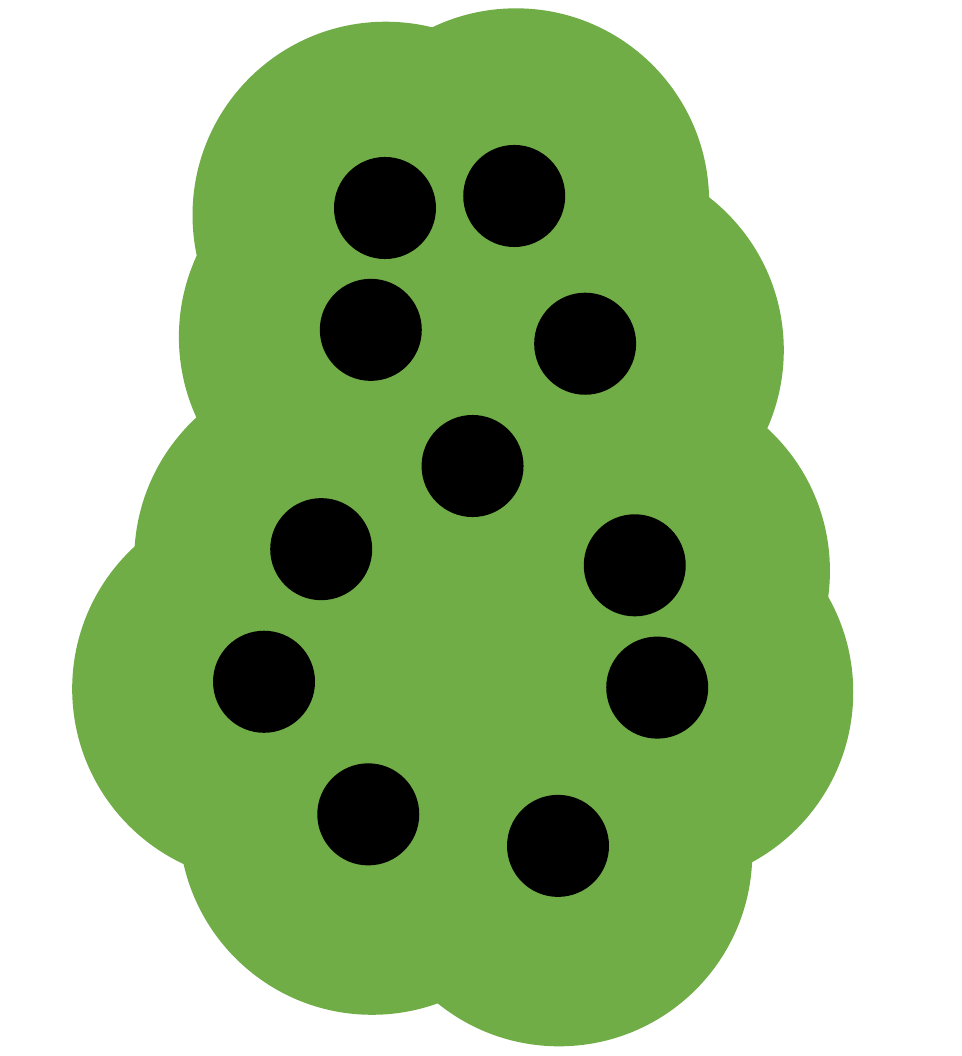} &
    \includegraphics[height=.1\textheight]{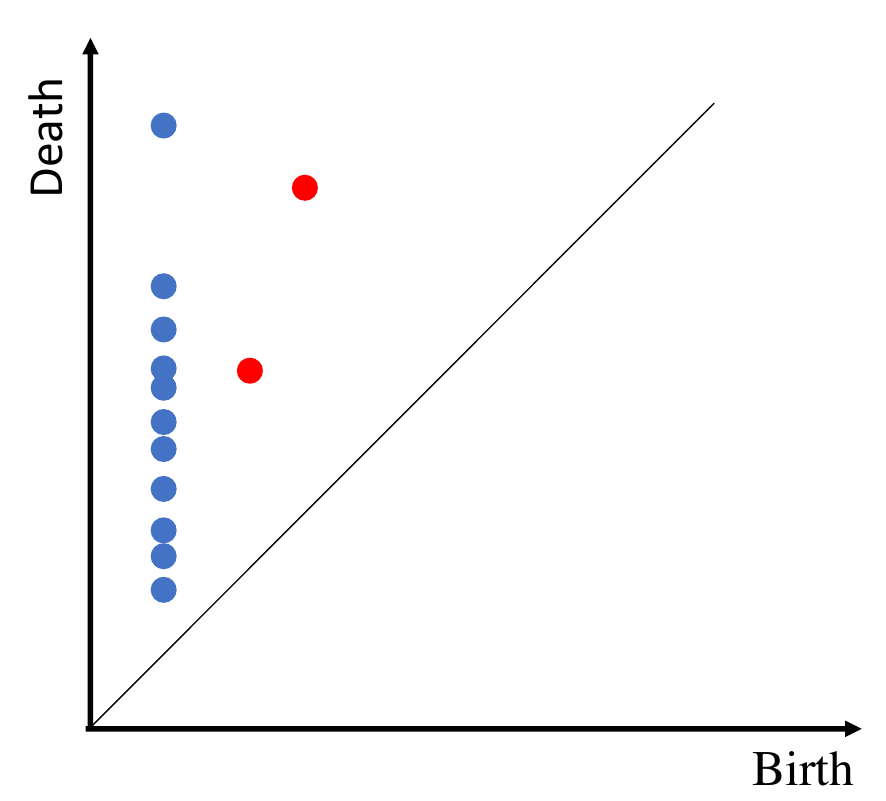} \\
        (a) & (b) & (c) & (d) & (e)
    \end{tabular}
 \caption{An illustration of persistence homology. \textbf{(a):} The change of topological structure while adapt a filtration on a set of points. The filtration value here is the Euclidean distance. \textbf{(b):} The corresponding PD. Blue points represent 0-dimensional homology, i.e., connected components. Red points represent 1-dimensional homology, i.e., loops. }
  \label{fig:PersistentHomology}
\end{figure}
	\section{Background: Persistence Homology}
	
	In this section, we introduce necessary background of persistent homology, distance between persistence diagrams, and the stability result.
	More technical details can be found in \cite{edelsbrunner2010computational,dey2022computational}.
	
	\myparagraph{Persistent homology} captures topological structures arising in data in a multi-scale and provably robust manner. The topological structures of interest are formalized in the language of algebraic topology \cite{munkers1984elements,hatcher2000algebraic}. Intuitively, they are connected components, loops, voids and their high-dimensional counterparts. Given a domain of interest, $X$, these topological structures are captured and measured through a \emph{filter function}, i.e., a scalar function defined on $X$, $f:X\rightarrow \mathbb{R}$. 

Intuitively, we filter the domain using the filter function and a continuously growing threshold, $\alpha$. As we increase $\alpha$, the part of the domain whose filter function value is below $\alpha$ will continuously grow.
Formally, we call the part below $\alpha$ a \emph{sublevel set}, \mbox{$X_a = \{x\in X|f(x) \leq a\}$}. 
As the threshold value $a$ increases from $-\infty$ to $\infty$, we obtain a sequence of growing spaces, called a \emph{filtration} of $X$: $\emptyset = X_{-\infty} \subset ... \subset X_{\infty} = X$. As $X_{\alpha}$ grows from $\emptyset$ to $X$, new topological structures gradually appear (born) and disappear (die). 
Fig.~\ref{fig:PersistentHomology}(a)-(d) show the growing process if we filter the space using a distance transform from given data.

Applying the homology functor to the filtration, we can more precisely quantify the birth and death of topological features (as captured by homology groups) throughout the filtration, and the output is the so-called \emph{persistence diagram (PD)}. A PD is a planar multiset of points, each of which corresponds to some homological feature (i.e., components, loops, and their higher dimensional analogs). The coordinates of a persistent point $p=(b,d)$ correspond to the birth and death times of the homological feature during the filtration. The lifetime of the feature, $\Pers(p)=|d - b|$, is called its \emph{persistence} and intuitively measures its importance w.r.t.~the input filter function. For technical reasons, we also add the diagonal line $\Delta$ to the diagram. Denote by $\dgm(f)$ the PD of a filter function $f$ regarding these homogolical features. 

\myparagraph{Computation and critical vertices.} In computation, we often decompose the domain into a discretization, e.g., a simplicial complex consisting of a set of simplices such as vertices, edges, triangles, etc. We assume the complex does not change over the course of optimization. The filter function $f$ is given at verices of the simplicial complex and is extended to all simplices. A simplex's filter function value is the maximum of its vertices, $f(\sigma) = \max_{v\in \sigma} f(v)$. With these filter function values, we construct the filtration as a sequence of subcomplexes (i.e., subsets of the complex). The computation of persistent homology boils down to a reduction algorithm on the boundary matrix encoding the combinatorial relationship between elements of the complex. Finally, we note that each persistent point corresponds to a pair of simplices and its birth/death times are the function values of these simplices, and eventually the function values of two vertices. Formally, for a persistent point $p\in \dgm(f)$, its birth and death times are $\birth(p) = f(v_b(p))$ and $\death(p) = f(v_d(p))$ for some vertices $v_b(p)$ and $v_d(p)$. We abuse the notation and call these vertices the \emph{birth and death vertices} of a persistent point $p$. This is crucial for the topological loss optimization; the gradient is essentially moving the persistent points by changing the function values of the relevant vertices $v_b(p)$ and $v_d(p)$. 

\myparagraph{Distance and stability of persistence diagrams.}
One of the most appealing traits of persistent homology is its stability, i.e., the distance between two diagrams is bounded by the difference between their input functions. 
A popular metric for PDs is the \emph{Wasserstein distance}. Given two tame functions $f,g: X\to \mathbb R$, let $\Pi:=\{\pi: \dgm(f)\to \dgm(g)\}$ denote the set of all bijections between their diagrams. The $q$-Wasserstein distance between $\dgm(f)$ and $\dgm(g)$ is 
\begin{equation}
    \label{eq:WassersteinDistance}
    \dist_{q}(\dgm(f),\dgm(g)) = \inf_{\pi\in\Pi}\left[\sum_{p\in \dgm(f)}(\birth(p)-\birth(\pi(p)))^q + (\death(p)-\death(\pi(p)))^q \right]^{1/q},
\end{equation}
in which $\birth(p)$ and $\death(p)$ are the two coordinates of the persistence point $p$. See Fig.~\ref{fig:matching}(a) for an illustration of the matching. We note the original distance metric for PDs, called the \emph{bottleneck distance}, is essentially the Wasserstein distance with $q=\infty$ \cite{cohen2007stability}.

The \emph{Wasserstein Stability Theorem} states that given two tame Lipschitz functions $f$ and $g$ defined on a triangulable compact metric space $X$, there exist constants $k$ and $C$ depending on $X$ and the Lipschitz constants of $f$ and $g$ so that for every $q \ge k$,
\begin{equation}
    \label{WassersteinStability}
    \dist_{q}(\dgm(f),\dgm(g))\le C\cdot ||f-g||_{\infty}^{1-k/q}.
\end{equation}
The first stability result was reported regarding the bottleneck distance \cite{cohen2007stability}. The stability for Wasserstein distance was proved in \cite{cohen2010lipschitz} and recently was improved in \cite{primoz2020wasserstein}.

	\section{Main Result: Convergence Rate of Regularized Topology-Aware Losses}

We start by introducing a generalized version of a regularized topology-aware loss. Previous loss functions typically include a standard supervision loss term and a topological loss based on Wasserstein distance between persistence diagrams of the model output and the ground truth. To obtain better optimization guarantees without sacrificing the efficacy, we make two modifications to the typical loss function: (1) we introduce a regulaizer term using total persistence; (2) we modify the standard topological loss. 
We conclude this section by introducing the optimization algorithm and stating the main theorem in an intuitive manner, namely, the regularized topological loss can be optimized efficiently with bounded convergence rate.

We first introduce a few notations for ease of exposition.
We are optimizing the learning model parameter $W$. We take the prediction of the model, $\phi_W(x)$, parameterised by $W$, and compare with ground truth $y$ for every datum $(x,y)\in \calD$. 
Meanwhile, the prediction $\phi_W$ induces a filter function, $f_W$. The topological property of the prediction is described by the corresponding PD, $\dgm(f_W)$. We will compare this diagram with a \emph{ground truth diagram}, $\orgtruedgm$, which describes the desired topology. 
If we have certain ground truth function, we can use it to compute $\orgtruedgm$. Otherwise, we can construct $\orgtruedgm$ using prior knowledge: assume the data has $\beta$ many topological structures, we can create $\orgtruedgm$ by putting $\beta$ many points in its upper-left corner (i.e., points with high persistence). In practice, we often assume the number of true structures is limited, i.e., $\beta$ is upper bounded by a small number $B$, which will be important in our theoretical bound.

A typical topology-aware loss in the literature involves a standard supervision loss term $L_{supv}$ and a topological loss term $L_{topo}$. The supervision loss term is task-dependent. It can be cross-entropy, mean squared error, neighborhood KL divergence, etc. The topological loss typically measures the similarity between the topology of the prediction and the desired topology. In order to obtain good convergence guarantee, we will change the typical formulation slightly. In addition, we will introduce a regularization term $L_{reg}$.

\begin{figure}[hbtp]
    \centering
    \begin{tabular}{c@{\hskip .7in}c@{\hskip .7in}c}
    \includegraphics[height=.12\textheight]{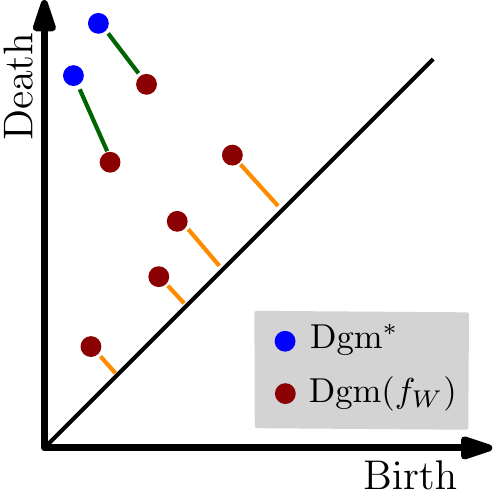} &
    \includegraphics[height=.12\textheight]{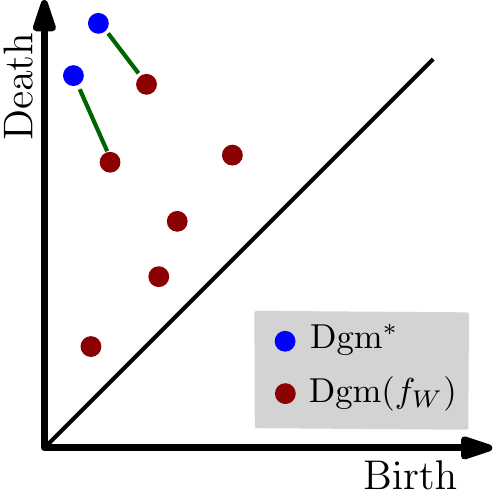} &
        \includegraphics[height=.12\textheight]{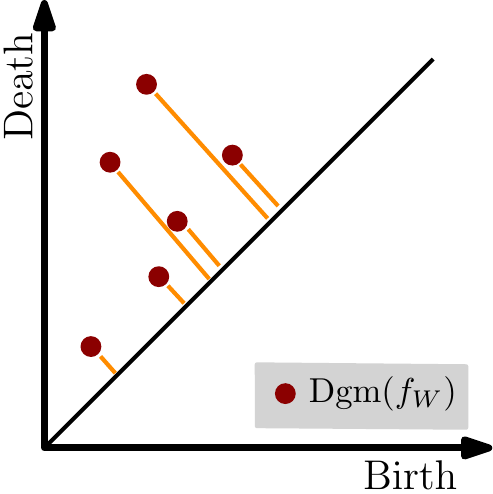} \\
        (a) & (b) & (c)
    \end{tabular}
    \caption{\textbf{(a):} Wasserstein distance between two diagrams, the ground truth diagram $\dgm^\ast$ and the diagram derived from the model prediction $\dgm(f_W)$. \textbf{(b):} the matching used for our topological loss term $L_{topo}$. All matchings between a non-diagonal point of $\dgm(f_W)$ and the diagonal from the Wasserstein distance (orange lines in Fig.~(a)) are ignored. \textbf{(c):} the total persistence is the cost of matching all non-diagonal points to the diagonal.}
    \label{fig:matching}
\end{figure}

\myparagraph{Regularization term.} We introduce a  regularization term using total persistence of the model output PD, $\dgm(f_W)$. The \emph{total persistence}, first introduced by Cohen-Steiner et al.~\cite{cohen2010lipschitz}, measures the ``norm'' of a single persistence diagram by aggregating the persistence of all its points. Formally, the $k$-th total persistence of $\dgm(f_W)$ is
\begin{equation}
    \label{eq:TotalPers}
    \TotPers_k(f_W) = \sum\nolimits_{p\in \dgm(f_W)} \Pers(p)^k = \sum\nolimits_{p\in \dgm(f_W)} \left[\death(p) - \birth(p)\right]^k 
\end{equation}
Note that the total persistence is essentially the matching distance between the diagram and an empty diagram; all non-diagonal points are matched to the diagonal of the empty diagram. See Fig.~\ref{fig:matching}(c) for an illustration. 
Optimizing this loss term will pull all diagram points toward diagonal, essentially ``shrinking'' all corresponding topological structures. Total persistence generally behave nicely w.r.t.~the input function (see Supplemental Material for details). It can play an important role in stabilizing the loss function, which is important for achieving our convergence bound.

\subsection{Topological loss, its modification and the rationale} 
Our topological loss term will be a modification of the typical topological loss.
The typical loss is the Wasserstein distance between the model output PD, $\dgm(f_W)$, and the ground truth PD, $\orgtruedgm$. 
As shown in Fig.~\ref{fig:matching}(a), it is the matching cost between $\dgm(f_W)$ (red points) and $\orgtruedgm$ (blue points). During the optimization, the gradient descent step will change the model weights, so as to change the output function $f_W$, and ultimately change the diagram. The red points will be moved toward their matches. Some red points are matched to non-diagonal blue points (highlighted with green lines). These are the structures we want to ``restore'' by increasing their persistence and moving them to match a salient structure in the ground truth. We call their cost the \emph{restoration cost}. The remaining red points are matched to the diagonal (highlighted with orange lines). The optimization will move these points toward the diagonal, essentially ``shrinking'' the corresponding structures, which are considered noise. We call their cost the \emph{shrinking cost}. 

To analyze the optimization behavior of this typical topological loss term is challenging. The main reason is that the loss is determined by a complex underlying combinatorial configuration. First, each persistent point corresponds to a pair of critical simplices. This correspondence and pairing can change as we update the model/function. Second, the loss is determined by an optimal matching between the diagrams. The matching may also change as we update the model/function. A gradient descent step will update function values according to a current configuration. But as a consequence, the step results in an updated configuration and thus an increased topological loss.

 To address this challenge, we need to control the increase of the loss function caused by the change of the underlying configuration after each gradient descent step. Recall the matching cost can be decomposed into restoration cost and shrinking cost. We can naturally decompose the topological loss into these two parts and treat them differently. We claim that the restoration cost can be reasonably bounded/controlled as it includes at most $B$ matchings. $B$ is the uppderbound of the number of true structures, and can be assumed reasonably small in practice. As for the shrinking cost, we observe that it is already contained within the total persistence/regularization term (Fig.~\ref{fig:matching})(c)). Thus, we can simply drop the shrinking cost from the topological loss and let the regularization term take care of shrinking these noise structures. 
\emph{This gives us the opportunity to bound the loss function.} In our theory, we will prove both the topological loss and the regularization term have a polynomial convergence rate. The convergence of the topological loss depends on the ground truth PD's cardinality bound, $B$. Whereas the convergence of $L_{reg}$ only depends on the volume of the domain, denoted by $C_X$.

With such consideration, we rewrite the topological loss term to only include the restoration cost. 

Formally, instead of matching all points between $\dgm(f_W)$ and $\dgm^\ast$, we ignore the diagonal line of $\dgm^\ast$, and only match its off-diagonal points to $\dgm(f_W)$ in an injective manner. 
See Fig.~\ref{fig:matching}(b) for an illustration. We denote by $\truedgm=\dgm^\ast\backslash\Delta$ the true diagram without the diagonal. The set of eligible injections from the true diagram to the prediction diagram is \mbox{$\Gamma(\dgm(f_W))=\{\gamma:\truedgm\to\dgm(f_W) \mid \gamma(p_1)\neq \gamma(p_2), \forall p_1\neq p_2\}$}. Our topological loss term is the optimal matching cost by any injection within $\Gamma$.
As for the power of the matching cost, similar to other previous works, we fix it to two and drop the power outside the summation\footnote{Extending to a general $p$-Wasserstain distance is not difficult and will be left to future work.}. 

 
Formally, our regularized topology-aware loss has three terms, $L_{supv}$, $L_{topo}$ and $L_{reg}$. 

\begin{definition}[Regularized Topology-Aware Loss]
\label{def:topoaware-loss}
A regularized topology-aware loss is 
\begin{eqnarray}
    G(W) &=& L_{supv}(W, \calD) + \lambda_{topo} L_{topo}(W, \calD) + \lambda_{reg}L_{reg}(W, \calD)\text{, in which}\\
    L_{supv}(W) &=& \sum\nolimits_{(x,y)\in \calD} \ell(\phi_W(x),y), \nonumber \\
    L_{topo}(W) &=& \min_{\gamma\in \Gamma}\sum\nolimits_{p\in \truedgm}\left[(\birth(p)-\birth(\gamma(p)))^2+(\death(p)-\death(\gamma(p)))^2\right], \nonumber \\
    L_{reg}(W) &=& \TotPers_k(f_W) \nonumber 
\label{eq:topoaware-loss}
\end{eqnarray}
Here $\ell(\cdot)$ is a point-wise supervision loss depending on the learning task. 
\end{definition}

	\subsection{Optimization Algorithm and Convergence Rate}
	Next, we introduce the optimization algorithm of the regularized topology-aware loss, and informally state our main theoretical result.

	Our algorithm iteratively updates the model weight and the corresponding persistence diagram accordingly. 
	Algorithm \ref{alg:optimization} shows our algorithm. It repeatedly recomputes the persistence diagram and the diagram matching, and updates the model weight via gradient descent. The gradient is dependent on the diagram and the matching. After the gradient decent step (i.e., after line 5 of the algorithm) the loss will decrease. But at the next iteration, after recomputing the diagram and the matching (lines 3 and 4), due to the change of the underlying configuration, the loss may increase again. The key of the theorem is to ensure the increasing due to the updating of the configuration is dominated by the decrease of the loss due to the gradient descent. For convenience, we denote by $f_t$ the function with parameter $W_t$, $f_t=f_{W_t}$. Its diagram $\dgm(f_t)$. And the corresponding optimal matching of the topological loss term is 
	$$\gamma_t = \argmax_{\gamma\in\Gamma(\dgm(f_t))} \sum_{p\in \truedgm}\left[(\birth(p) - \birth(\gamma(p)))^2 + (\death(p) - \death(\gamma(p)))^2\right]
	$$
	
	For better explanation, we introduce additional notations, $G_t$, $L_{supv}^t$, $L_{topo}^t$ and $L_{reg}^t$. They are the loss function and terms evaluated based on the underlying diagram at time $t$ and the matching at time $t$, $\gamma_t$. Note the underlying configuration may not match the parameter passed in. Recall that in a diagram, the coordinates of a persistent point $p$ is the function value of its birth and death vertices. We use the birth/death vertices in diagram at time $t$ for loss at time $t$, despite the input parameter. Therefore, we have 
	\begin{eqnarray}
	L^t_{topo}(W_t) =& \sum_{p\in \truedgm}\left[(\birth(p) - f_t(v_b(\gamma_t(p))))^2+(\death(p) - f_t(v_d(\gamma_t(p))))^2\right]\\
	L^t_{topo}(W_{t+1}) =& \sum_{p\in \truedgm}\left[(\birth(p) - f_{t+1}(v_b(\gamma_{t}(p))))^2+(\death(p) - f_{t+1}(v_d(\gamma_{t}(p))))^2\right]\\
	L^{t+1}_{topo}(W_{t+1}) =& \sum_{p\in \truedgm}\left[(\birth(p) - f_{t+1}(v_b(\gamma_{t+1}(p))))^2+(\death(p) - f_{t+1}(v_d(\gamma_{t+1}(p))))^2\right]
	\end{eqnarray}
	
	\begin{wrapfigure}{R}{0.5\textwidth}
	\vspace{-.2in}
\begin{minipage}{0.5\textwidth}
\begin{algorithm}[H]
\caption{Optimizing a Regularized Topology-Aware Loss}
\label{alg:optimization}
\textbf{Input:} $\calD$, $\truedgm$, learning rate $\eta$, convergence criterion $\epsilon$, and weights $\lambda_{topo}, \lambda_{reg}$. \newline
\textbf{Output:} Model weight $W$.

\begin{algorithmic}[1]
\STATE Randomly initialize $W_0$.
\FOR{$t=0,1,2,\ldots,T$}
    \STATE Compute $\dgm(f_{W_{t}})$ 
    \STATE Compute $\gamma_{t}$
    \STATE Gradient descent:\newline
        $W_{t+1} = W_{t} - \eta\nabla_W G_t(W_{t})$
    \IF{$|G_{t}(W_{t+1})-G_t(W_{t})|\le\epsilon$}
        \STATE Break the loop. Algorithm converges.
    \ENDIF
\ENDFOR
\RETURN $W_{t+1}$.
\end{algorithmic}
\end{algorithm}
\end{minipage}
	\vspace{-.2in}
 \end{wrapfigure}

	Note that the first two are evaluated using the same matched point $\gamma_t(p)$ and its birth/death vertices, but using different filter functions $f_t$ and $f_{t+1}$. The third loss is evaluated using the configuration at $t+1$ and using the new function $f_{t+1}$. Similarly, we can define $L^t_{reg}(W_t)$, $L^t_{reg}(W_{t+1})$, $L^{t+1}_{reg}(W_{t+1})$. We can also define $L^t_{supv}(W_t)$, $L^t_{supv}(W_{t+1})$, $L^{t+1}_{supv}(W_{t+1})$. But note $L^t_{supv}(W_{t+1})=L^{t+1}_{supv}(W_{t+1})$ as it does not involve the configuration update. Adding the terms together, we have the loss function $G_t=L^t_{supv}+L^t_{topo}+L^t_{reg}$.
	
	During the optimization, as we update the parameters, the loss function is updated as: \mbox{$\cdots G_t(W_t)\to G_t(W_{t+1})\to G_{t+1}(W_{t+1}) \cdots$}. The first arrow is when we calculate the gradient using the configuration at time $t$ and update the parameter to $W_{t+1}$ with gradient descent. During this step, the loss will monotonically decrease. The second arrow is when we keep the parameter unchanged. But update the underlying configuration with the new diagram $\dgm(f_{t+1})$, and the new optimal matching $\gamma_{t+1}$. In this step the loss may increase instead. In our theory, the main goal is to show that the loss decrease in step one is bigger than the loss increase in step two.

Finally, we state our main theorem in an informal manner. A complete version of the theorem and its proof will be provided in the next section. 
\begin{thm}[Main Theorem (Informal)]
Under mild regularity assumptions, with carefully chosen learning rate, $\eta$, Algorithm \ref{alg:optimization} stops (i.e., the change of $G$ is below $\epsilon$) in $O(1/\epsilon)$ iterations.
\end{thm}

The theorem states that the convergence rate of the algorithm is linear to $1/\epsilon$. This is considered very efficient. Note for non-convex optimization smooth functions, the $1/\epsilon^2$ convergence rate for reaching stationary point, i.e., $\|\nabla_W G_t(W_t)\|^2 \leq \varepsilon$ can be translated into a $1/\epsilon$ convergence rate using our stop criterion~\cite{nesterov2013introductory,jin2017escape}. 
	
	\section{A Complete Version of the Main Theorem}
In this section, we provide our main theorem on time complexity of the optimization algorithm. 
The $O(\cdot)$ notation applied here hides parameters which are free of $\epsilon$. To begin with, we first introduce following regularity assumptions of the behavior of the filter function $f_W$ and the supervision loss function $L_{supv}(W)$. Note these regularity conditions are standard in analyzing convergence of optimization algorithms and topological data analysis. 

	\begin{itemize}[topsep=1pt,itemsep=1pt,partopsep=1pt, parsep=1pt,leftmargin=8pt]
	\item \textbf{Assumpt.~1 (A1):} $f$ is  $1$-bounded, $1$-Lipschitz continuous and $1$-Lipschitz smooth relative to $W$. 
	
	\item \textbf{Assumpt.~2 (A2):} $L_{supv}(W)$ is $\ell^0$-bounded, $\ell^1$-Lipschitz continuous and $\ell^2$-Lipschitz smooth relative to $W$. 
	\end{itemize}

%
We show that by carefully choosing the step size/learning rate, one can control the process of optimization so that the bounce up of topological loss due to persistence diagram updating, $G_{t+1}(W_{t+1})-G_{t}(W_{t+1})$, will never exceed the overall loss reduction $G_{t}(W_{t})-G_{t}(W_{t+1})$. Indeed in each iteration, the total loss function is monotone decreasing thus will terminate efficiently. 
	\begin{thm} \label{thm_major}
		Assuming \textbf{A1} and \textbf{A2} hold, the algorithm stops in   $O\left(\frac{1}{\epsilon}\right)$  iterations if stepsize $\eta$ is chosen to be $\eta \leq \min\left\{\frac{1}{2(\ell^2+2\lambda_{reg} k (k+1)C_X+ 2\lambda_{topo} k{B})}, \frac{\sqrt{\varepsilon}}{1024\lambda^2_{topo}{B}^2 }, \frac{\sqrt{\varepsilon} }{ 16
    \lambda^2_{reg} k^2C^2_{ X}}\right\}$. Here ${B}$ is the cardinally of the ground truth diagram (without the diagonal), i.e., ${B}=\card(\truedgm)$. Constant $C_X$ is the volume of the domain and is free of $\varepsilon$.
	
		
	\end{thm} 
    The proof will need the following Lemmas. Some of their proofs will be left in Appendix.
        
  \begin{lemma}\label{lem_f4}
	Assume \textbf{A1} holds, we have
	\begin{enumerate}
	    \item $\lambda_{topo}L_{topo}(W_{t})+\lambda_{reg}L_{reg}(W_{t}) \le \lambda_{topo}{B} + \lambda_{reg}C_{X}$. 
	    \item $\|\nabla_{W}\lambda_{topo}L_{topo}(W_{t})+\nabla_{W} \lambda_{reg}L_{reg}(W_{t})\|_2\le 2\lambda_{reg} k C^2_X+ 2\lambda_{topo} k{B}$. 
	    \item $\|\nabla_{W}^2\lambda_{topo}L_{topo}(W_{t})+\nabla_{W}^2 \lambda_{reg}L_{reg}(W_{t})\|_2\le 2\lambda_{reg} k (k+1)C_X+ 2\lambda_{topo} k{B}$.
	\end{enumerate}
\end{lemma}
    
    With Lemma~\ref{lem_f4} and Assumption {A2} we can prove following facts, which bounds zero-th order, first order and second order derivative of  $G_t(W_t)$: 


		

		\textbf{Fact 1: Bounded function value}:  $G_t(W_t)\leq\ell^0 + \lambda_{reg} C_X + \lambda_{topo} {B} \overset{\Delta}=C_0$ 
		
		\textbf{Fact 2: Bounded gradient}: $\|\nabla_{W}G_t(W_t)\|_2\leq \ell^1+2\lambda_{reg} k C_X+ 2\lambda_{topo} k{B}\overset{\Delta}{=}C_1$

		\textbf{Fact 3: Bounded  Hessian}: $\|\nabla^2_{W}G_t(W_t)\|_2\leq \ell^2+2\lambda_{reg} k (k+1)C_X+ 2\lambda_{topo} k{B} \overset{\Delta}{=} C_2$

For convenience, we denote these bounds by $C_0$, $C_1$ and $C_2$, respectively. Based on this lemma, we can show the following lemma.

\begin{lemma} [Improve or Localize ~\cite{jin2021nonconvex}] 
The parameter changing magnitude is controlled by the step size and the loss change.
$\|W_{t+1}-W_{t}\| \leq 2\sqrt{\eta(G_t(W_t)-G_t(W_{t+1}))}$.
\label{lem:improve-or-local}
\vspace{-0.2in}
\end{lemma}
\begin{proof}
	By \cite{nesterov2013introductory,jin2021nonconvex} we have $G_t(W_{t+1})\leq G_t(W_t) + \nabla G_t(W_t) ^\top [W_{t+1}-W_{t}]+ \frac{C_2}{2}\|W_{t+1}-W_t\|^2$. 
	Using the update equation $W_{t+1}=W_{t} -\eta \nabla_W G_t(W_{t})$, and pick $\eta \leq \frac{1}{2C_2}$ we have the following inequality:
	$G_t(W_t)-G_t(W_{t+1})\geq \frac{\eta}{4}\|\nabla_W G_t(W_t)\|_2^2=\frac{1}{4\eta}\|W_{t+1}-W_{t}\|^2$.
This inequality implies the lemma.
\vspace{-0.1in}
\end{proof}

\begin{lemma}[Bounded Increase of the Regularization Term]
The increase of the regularization term is bounded: $|L^t_{reg}(W_t)-L^{t+1}_{reg}(W_{t+1})|\leq 2kC_{ X}\sqrt{\eta (G_t(W_t)-G_t(W_{t+1}))}$.
\vspace{-0.2in}
\end{lemma}
\begin{proof}
By Lemma \ref{lem:improve-or-local} and assumption \textbf{A1}, $\|f_{t}-f_{{t+1}}\|_{\infty}\leq 2\sqrt{\eta(G_t(W_t)-G_t(W_{t+1}))} $.  After updating $\dgm(f_{{t+1}})$ and $\gamma_{t+1}$, because of the Lipschitz condition of total persistence (Lemma 8 
in the supplemental), $|\TotPers_k(f_{t})-\TotPers_k(f_{{t+1}})|\leq 2kC_{ X}\sqrt{\eta (G_t(W_t)-G_t(W_{t+1}))} $.
\vspace{-0.1in}
\end{proof}	
 \begin{lemma}[Bounded Increase of the Topological Term]
The increase of the topological term due to configuration change is bounded: $|L^t_{topo}(W_{t+1})-L^{t+1}_{topo}(W_{t+1})|\leq 16B\sqrt{\eta (G_t(W_t)-G_t(W_{t+1}))}$.
\vspace{-0.2in}
\end{lemma}
\begin{proof}
    Let $\gamma_{t\to t+1}$ be the optimal bijection between $\dgm(f_{t})$ and $\dgm(f_{t+1})$, i.e., the one corresponding to the Wasserstein distance.
    By the main theorem in \cite{cohen2007stability}, we have $\forall p\in \dgm(f_{t})$,
    \begin{equation}
        \begin{aligned}
            &\max \{|\birth(p)-\birth(\gamma_{t\to t+1} (p))|,|\death(p)-\death(\gamma_{t\to t+1} (p))|\}\\
            \leq & \|f_{t}-f_{t+1}\|_{\infty}\leq 2\sqrt{\eta(G_t(W_t)-G_t(W_{t+1}))}
        \end{aligned}
    \end{equation}
    The composition mapping $ \gamma_{t\to t+1}\circ \gamma_{t} (\cdot) $ is an injection from $\truedgm$ to $\dgm(f_{{t+1}})$. We have:
    
    \begin{equation*}
        \begin{aligned}
        &L^t_{topo}(W_{t+1})
            = \sum_{p \in \truedgm}| \birth(p)-f_{t+1}(v_b(\gamma_{t}(p)))|^2+|\death(p)-f_{t+1}(v_d(\gamma^*_{t}(p)))|^2\\
            = & \sum_{ p \in \truedgm}| \birth(p)-f_{t+1}(v_b(\gamma_{t\to t+1}(\gamma_{t} (p))))+ f_{t+1}(v_b(\gamma_{t\to t+1}(\gamma_{t} (p))))-f_{t+1}(v_b(\gamma_{t}(p)))|^2\\
            &+|\death(p)-f_{t+1}(v_d(\gamma_{t\to t+1}(\gamma_{t} (p))))+f_{t+1}(v_d(\gamma_{t\to t+1}(\gamma^*_{t} (p)))) -f_{t+1}(v_d(\gamma_{t}(p)))|^2\\
            \geq&  \sum_{ p \in \truedgm}| \birth(p)-f_{t+1}(v_b(\gamma_{t\to t+1}(\gamma_{t} (p))))|^2 +|\death(p)-f_{t+1}(v_d(\gamma_{t\to t+1}(\gamma_{t} (p))))|^2
        \end{aligned}
    \end{equation*}
    \begin{equation}
        \begin{aligned}
    &- 2|\birth(p)-f_{t+1}(v_b(\gamma_{t\to t+1}(\gamma_{t} (p))))||f_{t+1}(v_b(\gamma_{t\to t+1}(\gamma_{t} (p))))-f_{t+1}(v_b(\gamma_{t}(p)))|\\
    &-2|\death(p)-f_{t+1}(v_d(\gamma_{t\to t+1}(\gamma_{t} (p))))||f_{t+1}(v_d(\gamma_{t\to t+1}(\gamma_{t} (p))))-f_{t+1}(v_d(\gamma_{t}(p)))|\\
     \geq&  \sum_{ p \in \truedgm}| \birth(p)-f_{t+1}(v_b(\gamma_{t\to t+1}(\gamma_{t} (p))))|^2 - 16\sqrt{\eta(G_t(W_t)-G_t(W_{t+1}))}\\
    &+|\death(p)-f_{t+1}(v_d(\gamma_{t\to t+1}(\gamma_{t} (p))))|^2 - 16\sqrt{\eta(G_t(W_t)-G_t(W_{t+1}))} \\
    \geq & -32\sqrt{\eta(G_t(W_t)-G_t(W_{t+1}))} \\
    +&  \sum_{ p \in \truedgm}| \birth(p)-f_{t+1}(v_b(\gamma_{t+1} (p)))|^2 +|\death(p)-f_{t+1}(v_d(\gamma_{t+1} (p)))|^2\\
    = & -32\sqrt{\eta(G_t(W_t)-G_t(W_{t+1}))} + L^{t+1}_{topo}(W_{t+1}) 
    \end{aligned}
\end{equation}
    The last but two inequality holds because $|f_{t+1}(v_b(\gamma_{t\to t+1}(\gamma_{t} (p))))-f_{t+1}(v_b(\gamma_{t}(p)))|\le 2\|f_t-f_{t+1}\|_\infty \le 4\sqrt{\eta(G_t(W_t)-G_t(W_{t+1}))}$, and $| \birth(p)-f_{t+1}(v_b(\gamma_{t\to t+1}(\gamma_{t} (p))))| \leq 2$ from \textbf{A1}.
    The last inequality is due to the fact that $\gamma_{t\to t+1}(\gamma_{t} (p)) \in \dgm(f_{{t+1}})$ and the optimality of $\gamma_{t+1}$.
    \vspace{-0.1in}
\end{proof}    
Finally, we are ready to prove our main theorem. We have the decrease of loss at each iteration:
    \begin{align}
      &G_t(W_{t})-G_{t+1}(W_{t+1}) \nonumber\\
      =&G_t(W_{t})-G_t(W_{t+1})+G_t(W_{t+1})-G_{t+1}(W_{t+1}) \nonumber\\
      \geq & G_t(W_{t})-G_t(W_{t+1})  - \left(32\lambda_{topo}{B} + 2 \lambda_{reg}kC_{ X} \right)\sqrt{\eta(G_t(W_t)-G_t(W_{t+1}))} 
    \end{align}
    By the algorithm, before the optimization stops, $G_t(W_{t})-G_{t}(W_{t+1})\geq \varepsilon$. 
    To ensure that $G_t(W_{t})-G_{t+1}(W_{t+1}) \geq \frac{G_t(W_{t})-G_{t}(W_{t+1}) }{2} \geq \frac{\epsilon}{2}$ we need an $\eta$ that satisfies  $\eta \leq  \min\left\{ \frac{\sqrt{ (G_t(W_t)-G_t(W_{t+1}))}}{4096\lambda^2_{topo}{B}^2 }, \frac{\sqrt{ (G_t(W_t)-G_t(W_{t+1}))} }{ 16
    \lambda^2_{reg} k^2C^2_{ X}}\right\}$.
    Leveraging the fact that $(G_t(W_t)-G_t(W_{t+1})) \geq \varepsilon$ and combining the constraint on $\eta \leq \frac{1}{2C_2}$, it suffices to pick $\eta\leq  \min\left\{\frac{1}{2(\ell^2+2\lambda_{reg} k (k+1)C_X+ 2\lambda_{topo} k{B})}, \frac{\sqrt{\varepsilon}}{1024\lambda^2_{topo}{B}^2 }, \frac{\sqrt{\varepsilon} }{ 16
    \lambda^2_{reg} k^2C^2_{X}}\right\}$, 
 we have $G_t(W_{t})-G_{t+1}(W_{t+1})\geq \frac{\epsilon}{2}$ if the stop criterion is not satisfied. Thus the algorithm stops in $\frac{2C_0}{\epsilon}$ iterations.
 
 This concludes the proof.

\begin{remark}
Theorem ~\ref{thm_major} suggests that by carefully choosing step-size, the optimization procedure of the topology-aware loss function terminates efficiently. When proper step size is taken, the loss function will decrease monotonically, thus potential oscillation due to the updating of configuration is suppressed. Note that there are three terms controlling the choice of the learning rate, i.e., $\frac{1}{2(\ell^2+2\lambda_{reg} k (k+1)C_X+ 2\lambda_{topo} k{B})}$, $\frac{\sqrt{\varepsilon}}{1024\lambda^2_{topo}{B}^2 }$ and $\frac{\sqrt{\varepsilon} }{ 16\lambda^2_{reg} k^2C^2_{X}}$. The first two depends on the the cardinally of $\truedgm$. Our prior assumption that $B$ is small is crucial here to ensure the choice of the step size does not have to be too small; if $B$ is linear to the dataset, then the step-size will be impossible to choose in practice.
\end{remark}

\section{Experiment}


\noindent\begin{minipage}{\linewidth}
\centering
  \begin{tabular}{cc}
  \includegraphics[height=.2\textheight]{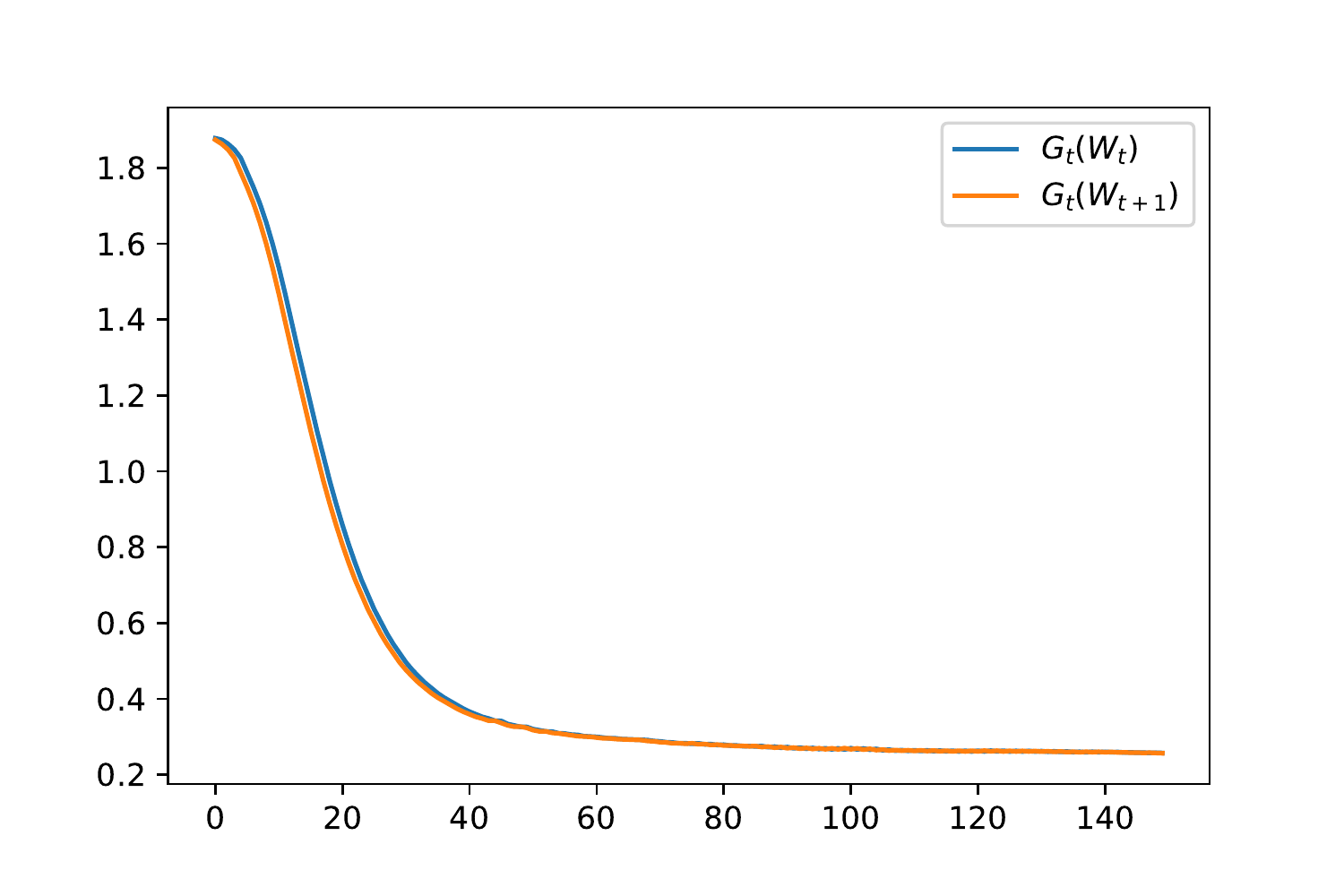} &
    \includegraphics[height=.2\textheight]{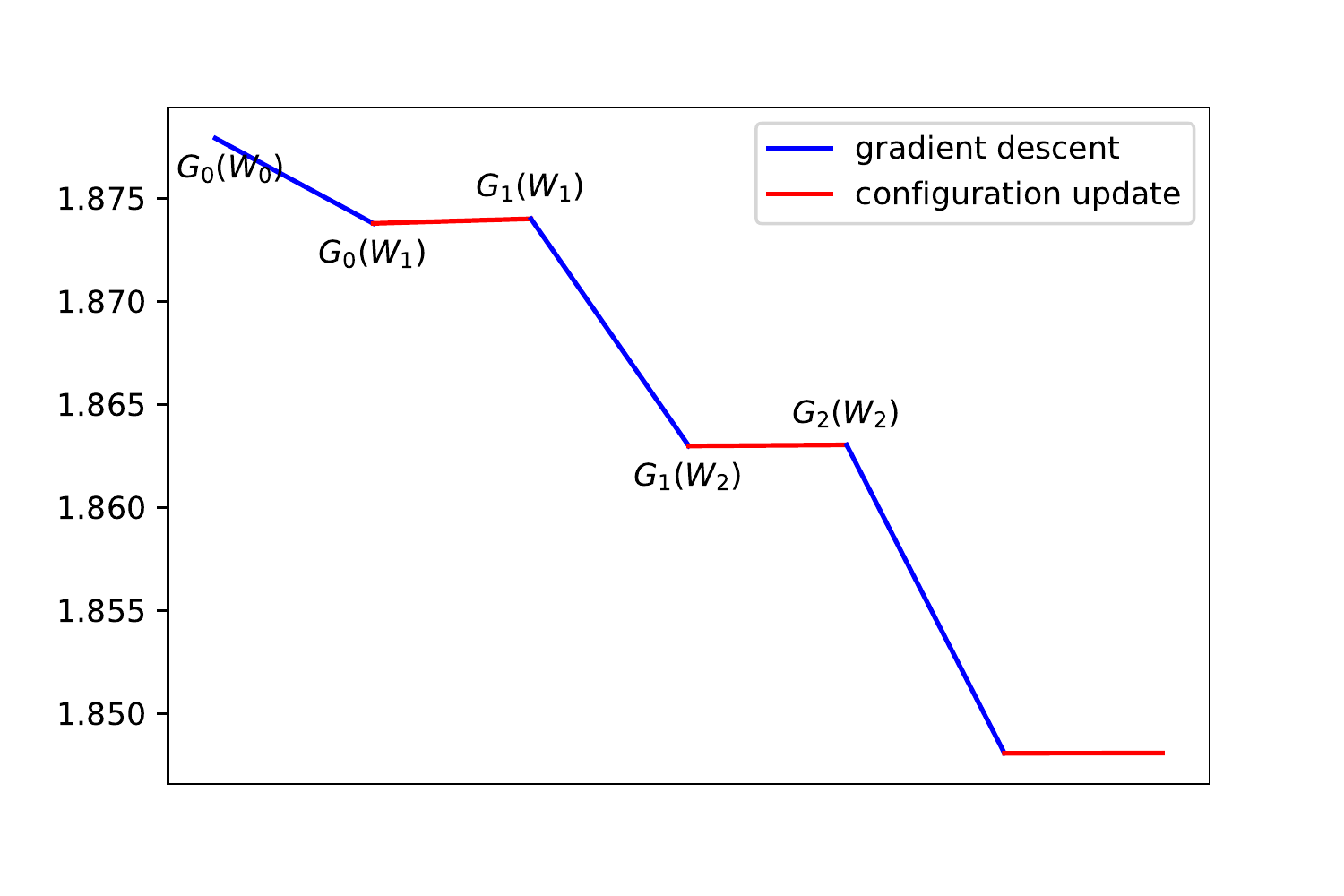} \\
     (a) & (b)
    \end{tabular}
  \captionof{figure}{\textbf{(a):} The optimization of $G_t(W_t)$ and $G_t(W_{t+1})$. \textbf{(b):} The loss decrease by gradient decent ($G_t(W_t)\to G_t(W_{t+1})$) is larger than the loss increase by configuration update ($G_t(W_{t+1})\to G_{t+1}(W_{t+1})$). }
  \label{fig:update}
 %

%
%
     \captionof{table}{The value of $G_t(W_t)$, $G_t(W_{t+1})$, and $G_{t+1}(W_{t+1})$ for the first few iterations.}
        \begin{tabular}{cccc}
            \toprule
            $t$ & 0 & 1 & 2 \\
            \hline
            $G_t(W_t)$ & 1.87793 & 1.87402 & 1.86304\\
            $G_t(W_{t+1})$ & 1.87380 & 1.86299 & 1.84806 \\
            $G_{t+1}(W_{t+1})$ & 1.87402 & 1.86304& 1.84807 \\
            \bottomrule
        \end{tabular}
        \label{tab:update}
\end{minipage}


We show how our algorithm can be used in a dimension reduction problem. In this task, we embed high-dimensional data to low-dimensional space while keeping their topological properties. Here we denote the high dimension data and its embedding by $X, Y$. \textit{t-SNE} is a popular dimension reduction method. It minimizes the Kullback–Leibler divergence between the high-dimensional data distribution $P(X)$ and the embedded low-dimensional data distribution $Q(Y)$, i.e.,
\begin{equation}
\label{eq:EmbeddingLoss}
    L_{supv}(X, Y) = \sum_{i\neq j} P_{ij}\log(P_{ij}/Q_{ij}).
\end{equation}
Here, the distribution $P_{ij}=(P_{i|j}+P_{j|i})/2N$. $P_{i|j}$ is calculated by
$P_{i|j} = \frac{\exp{(-|x_i-x_j|^2/2\sigma_i^2)}}{\sum_{k\neq i}\exp{(-|x_i-x_k|^2/2\sigma_i^2)}}$, and the distribution $Q_{ij} = \frac{(1+|y_i-y_j|^2)^{-1}}{\sum_{k}\sum_{l\neq k}(1+|y_k-y_l|^2)^{-1}}$. 

We feed the high-dimensional point cloud to a neural network with four fully-connected layers and embed it into lower dimension space. In this case, the true PD is calculated from the original point cloud $X$ and the target PD is calculated from the embedded point cloud $Y$. To reserve topological features during optimization, the topological difference $L_{topo}$ and regularization $L_{reg}$ is added to \eqref{eq:EmbeddingLoss}, leading to a neighbor embedding topology-aware loss.

\begin{figure}[hbt]
\centering
  \begin{tabular}{c@{\hskip -0.05in}c@{\hskip -0.2in}c@{\hskip -0.15in}c}
  \includegraphics[height=.11\textheight]{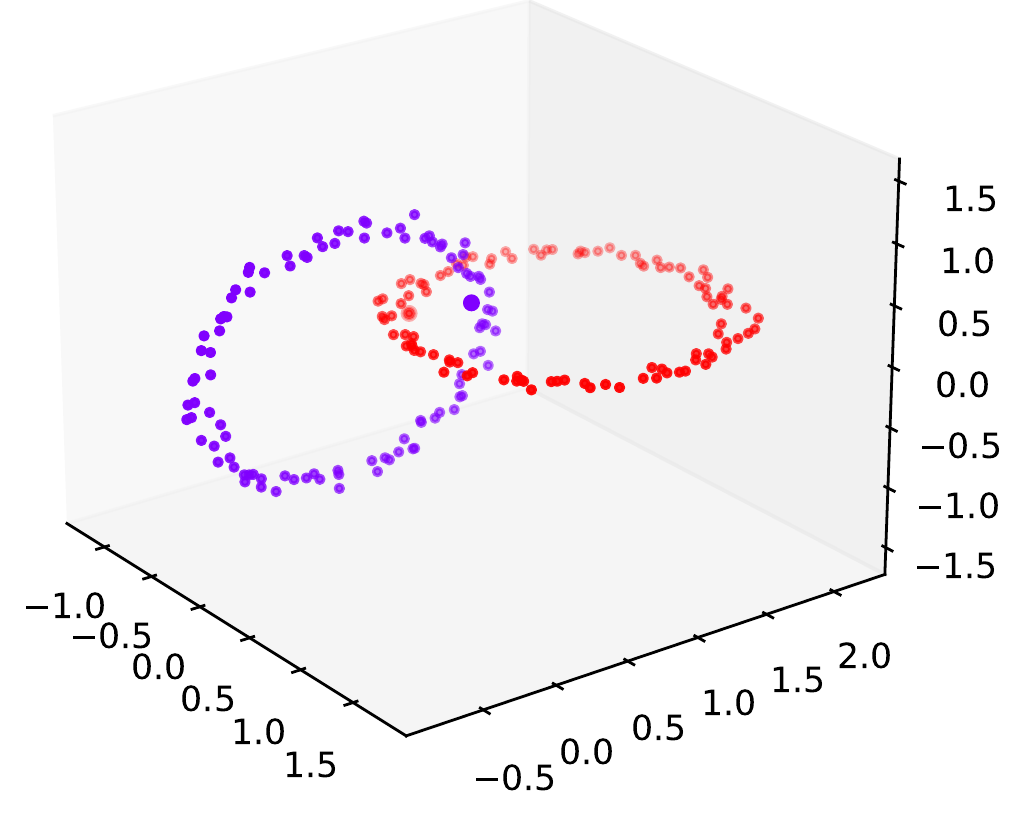} &
    \includegraphics[height=.11\textheight]{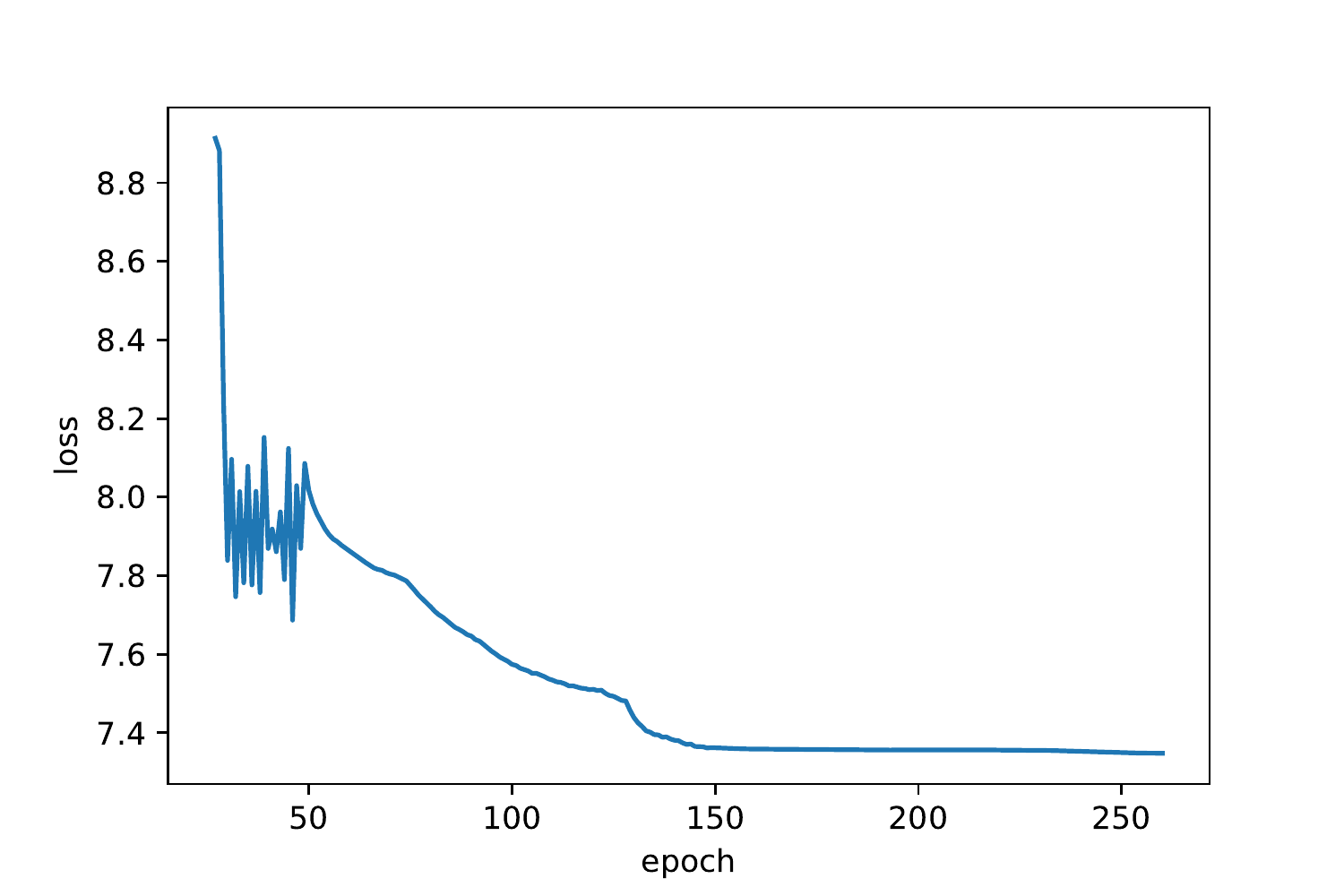} &
    \includegraphics[height=.11\textheight]{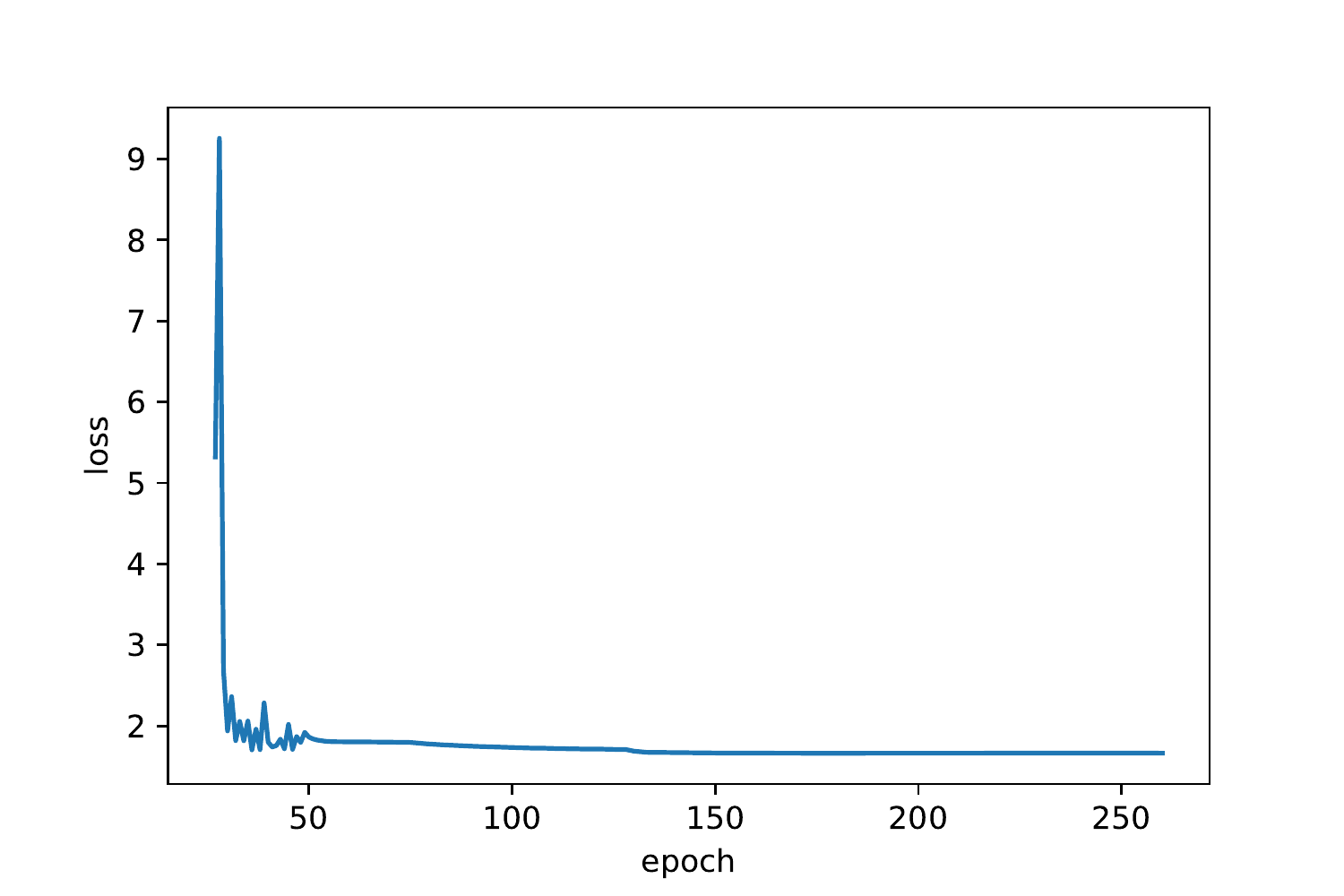} &
    \includegraphics[height=.11\textheight]{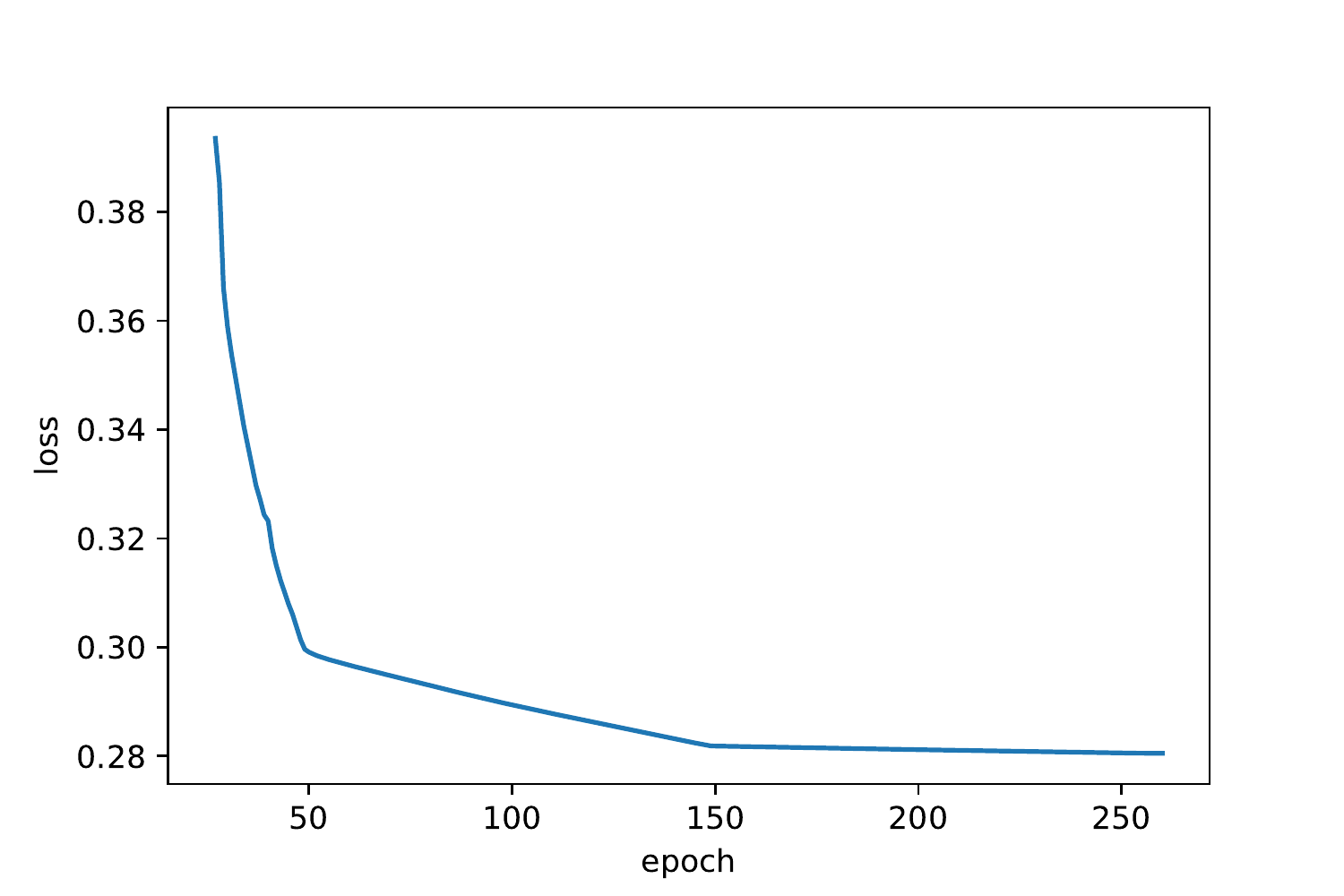} \\
     \tiny{(a) Original point cloud} & \tiny{(b) $L_{topo}$} & \tiny{(c) $L_{reg}$} & \tiny{(d) $G(W)$}
    \end{tabular}
  \caption{\textbf{(a):} The original point cloud for embedding. \textbf{(b)$\to$(d):} The convergence behaviour of $L_{topo}$, $L_{reg}$ and $G(W)$ during optimization. }
  \label{fig:convergence}
\end{figure}

\begin{figure}[hbt]
\centering
  \begin{tabular}{c@{\hskip -0.1in}c@{\hskip -0.1in}c@{\hskip -0.1in}c}
    \includegraphics[height=.1\textheight]{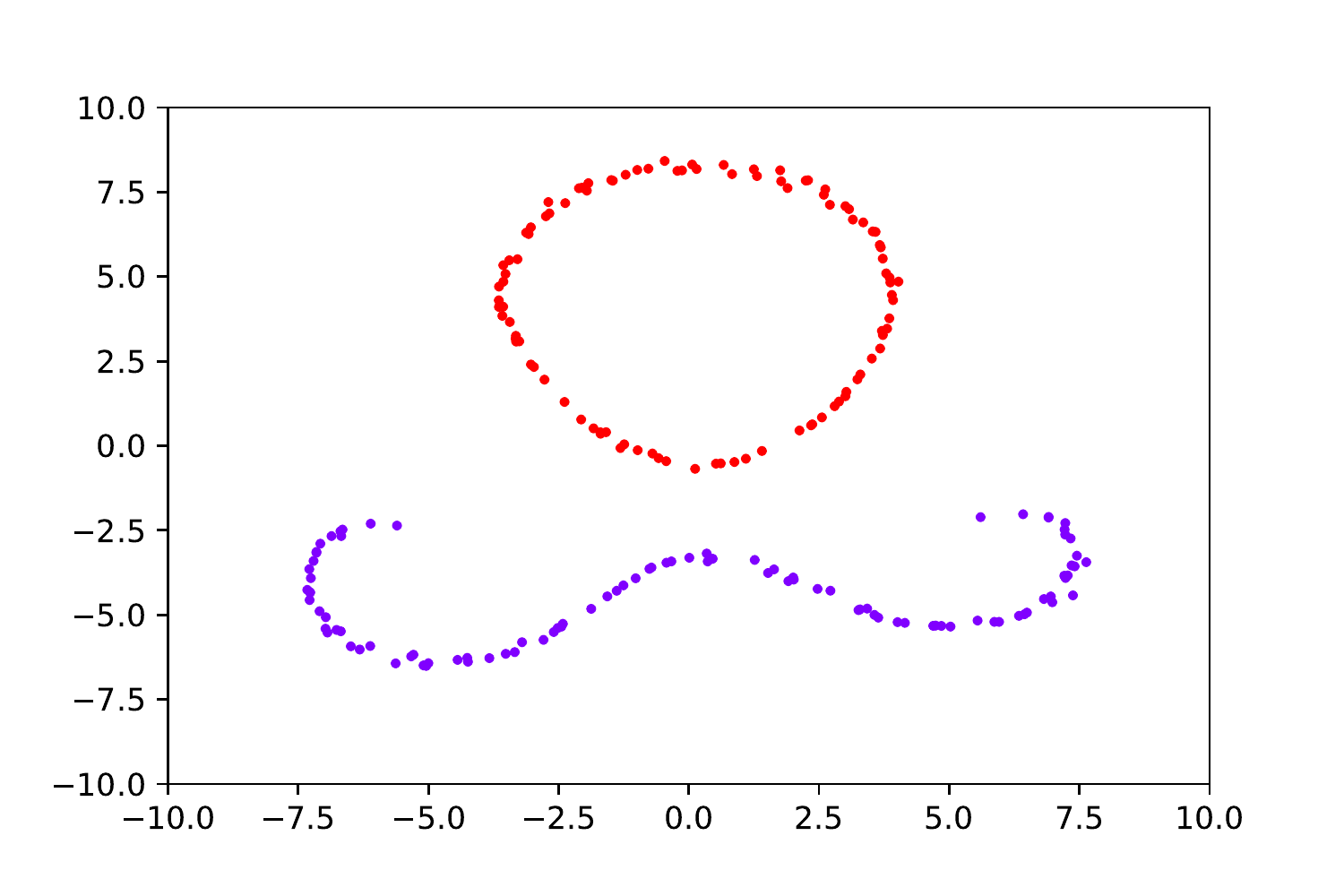} &
    \includegraphics[height=.1\textheight]{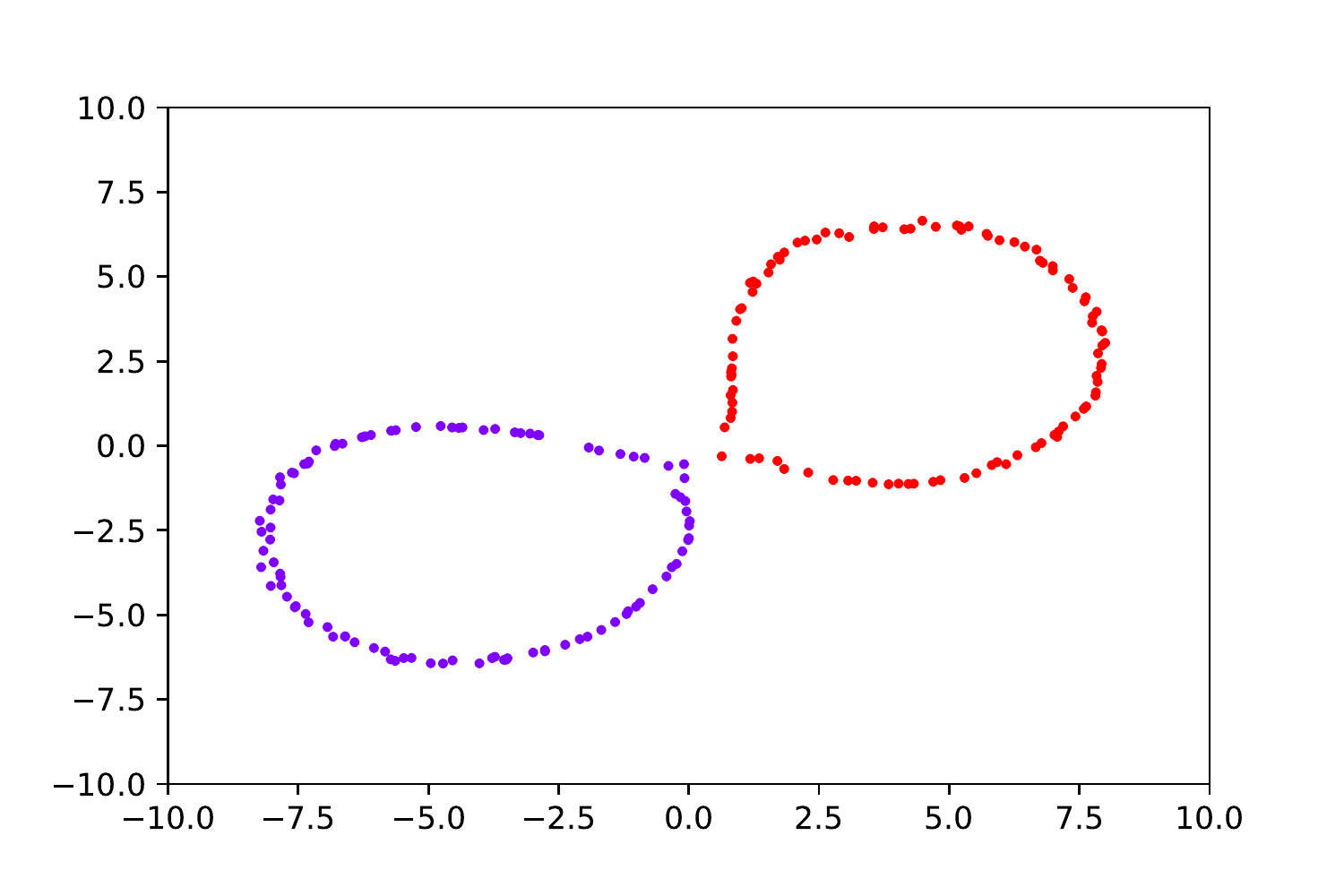} &
    \includegraphics[height=.1\textheight]{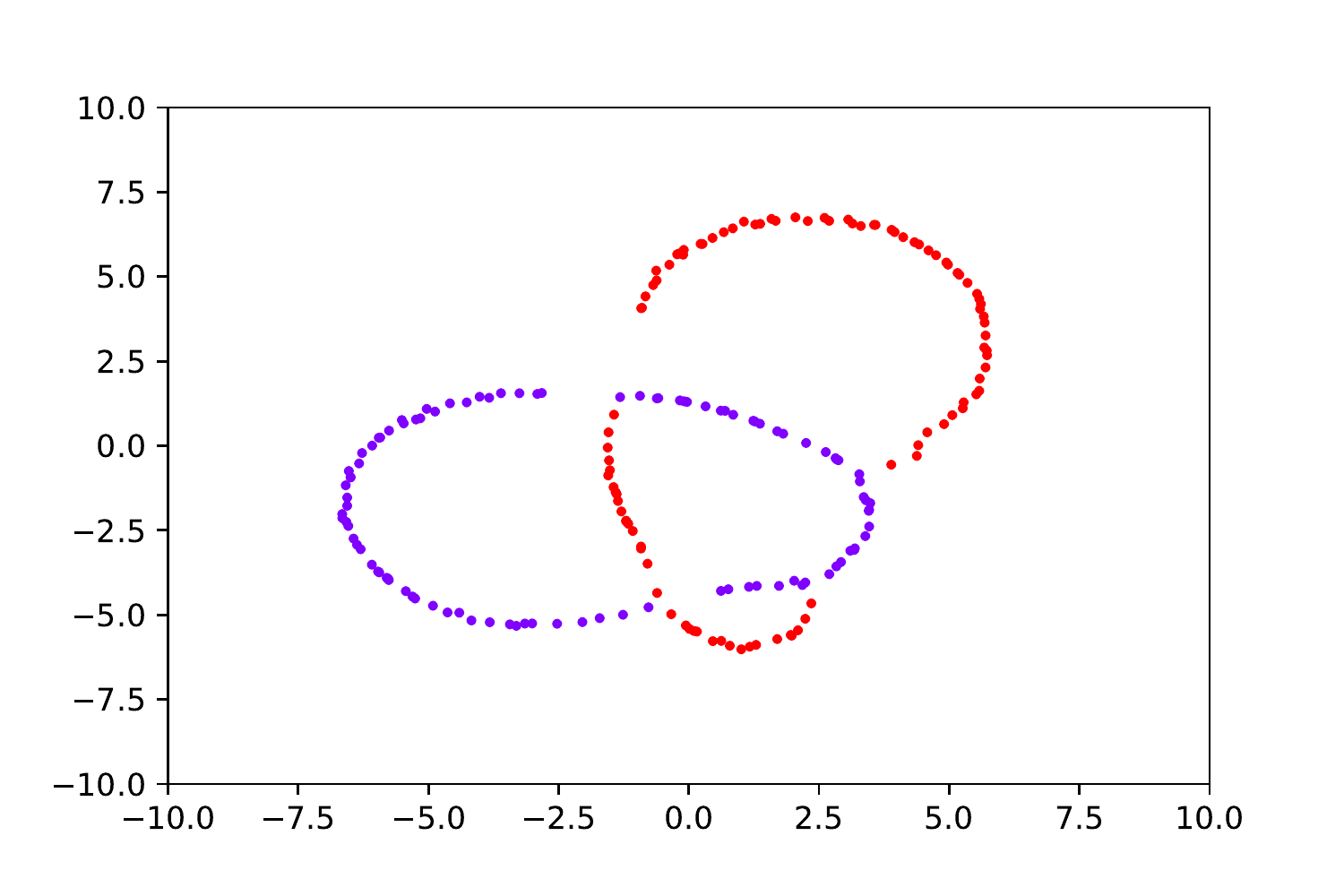} &
    \includegraphics[height=.1\textheight]{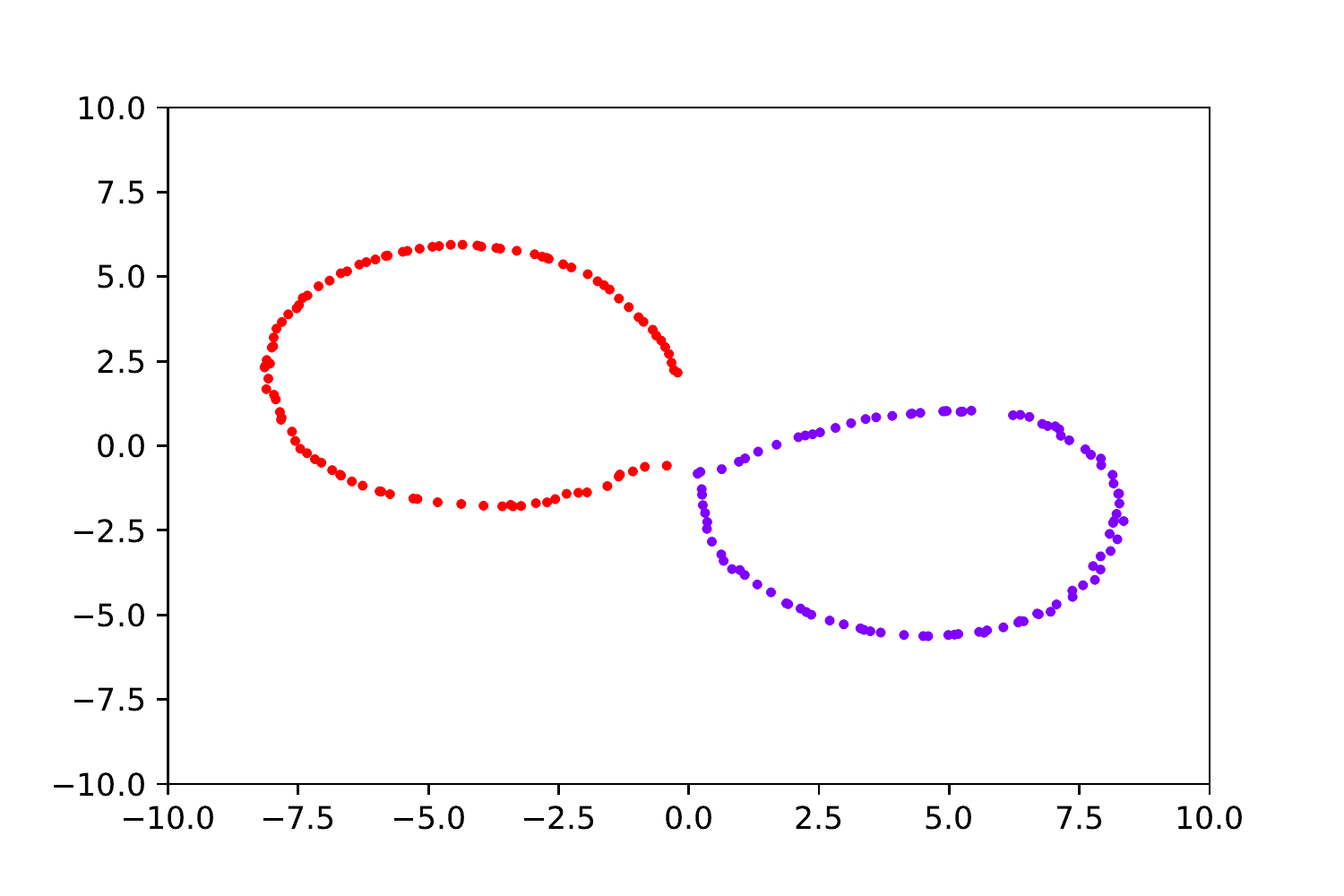} \\
        \tiny{(a) $\begin{array}{l}
             \lambda_{topo}=0 \\
             \lambda_{reg} =0
        \end{array}$ } & \tiny{(b) $\begin{array}{l}
             \lambda_{topo}=0.0005 \\
             \lambda_{reg} =0.005
        \end{array}$} & \tiny{(c) $\begin{array}{l}
             \lambda_{topo}=0.0005 \\
             \lambda_{reg} =0.01
        \end{array}$} & \tiny{(d) $\begin{array}{l}
             \lambda_{topo}=0.005 \\
             \lambda_{reg} =0.01
        \end{array}$}
    \end{tabular}
  \caption{Ablation study of $L_{topo}$ and $L_{reg}$. The $\mathbb R^3$ point cloud in Fig.\ref{fig:convergence} is embedded into $\mathbb R^2$. \textbf{(a)} The blue circle is degenerated into a curve without topological loss. \textbf{(b)} With proper $\lambda_{topo}$ and $\lambda_{reg}$, two circles are reserved and separated well. \textbf{(c)} But if $\lambda_{reg}$ is high, the two circles will be blended. \textbf{(d)} When we keep the high value of $\lambda_{reg}$ and increase $\lambda_{topo}$, two circles are separated again.}
  \label{fig:ablation}
\end{figure}

Following \cite{carriere2021optimizing}, We use a point cloud in $\mathbb R^3$ that is comprised of two nested circles, as shown in Fig.~\ref{fig:convergence} (a). We then embed it into $\mathbb R^2$. The persistent diagram is calculated only with 0-dimensional homology. 
In Fig.~\ref{fig:ablation}, we show the embedded point cloud without topological loss (a) and with topological loss (b)-(d). Results show that the regularized topology-aware loss can successfully preserve the blue circle.

\myparagraph{Numerical Illustration of the Two-step Optimization Process.}
We provide experimental illustration supporting our theoretical insights.
Each optimization iteration involves two steps: update the model with gradient descent; and recalculate the topological configuration (persistence diagram and matching).
With the same topological configuration, the gradient descent can decrease the loss function $G_t(W_t)\to G_t(W_{t+1})$. However, when updating the configuration, the loss changes from $G_t(W_{t+1})$ to $G_{t+1}(W_{t+1})$. This update may increase the loss function instead. Our main theory proves the loss increase brought by configuration update is smaller than the loss decrease by gradient descent.  We provide empirical evidence to validate this.
We follow the experimental setting in our main paper. In Fig.~\ref{fig:update} (a) we show that both $G_t(W_t)$ and $G_t(W_{t+1})$ are monotonically decreasing and converge to the same value, when their difference is less than $\epsilon$. In Fig.~\ref{fig:update} (b) and Table \ref{tab:update}, we show that the configuration update can actually raise the loss value, but the gradient decent lowers it much more. Thus the total loss $G_t(W_t)$ can decrease monotonically.

\myparagraph{Convergence.} 
In Fig.~\ref{fig:convergence}, we show the loss plot over the training process. We observe oscillation of both $L_{topo}$ and $L_{reg}$, which is expected and is due to the change of the configuration. But the total loss function $G(W)$ is monotonically decreasing, as we proved in the theorem.

\myparagraph{Ablation on $L_{topo}$ and $L_{reg}$.} The ablation study of $L_{topo}$ and $L_{reg}$ in Fig.~\ref{fig:ablation} illustrates their effect in practice. In (a), we show the case without topological loss. The topology cannot be preserved. In (b), with a reasonable amount of topological loss and the regularization, we preserve the topology as desired. In (c), we increase the weight of the regularization term. This leads to stronger force to shrink all data into a point. In (d), we increase the weight of the topological loss. It balances the contracting force of the regularization term, and better preserves the topology.

\section{Conclusion}
In this paper, we investigated the optimization behavior of topology-aware loss. By introducing a regularization term based on total persistence, and modifying the typical topological loss term, we provide a general version regularized topology-aware loss function. The main result of the paper is theoretical; we show that the loss can be efficiently optimized with a carefully chosen learning rate. 

\newpage
\bibliographystyle{plain}
\bibliography{topo}

\newpage
\appendix
\section*{Appendix}
\section{Analysis on Total Persistence Moments}
\label{sec:prelim}
To analyze the Lipschitz stability of total persistence, we need extra definitions. The following definitions and lemmas are from \cite{cohen2010lipschitz}.
	\begin{definition} [Mesh of Triangulation]
		Let $X$ be a triangulable, compact metric space and $\rm{dist} (\cdot,\cdot): X \times X \rightarrow \mathbb{R}$ is the metric. Given a triangulation $\mathcal{K}$ of $X$, the \textit{diameter} of simplex $\Delta$ is denoted as $\rm{diam}(\Delta)= \max\limits_{x,y \in \Delta} \rm{dist}(x,y)$. The \textit{mesh} of a complex $\mathcal{K}$ is $\rm{mesh}(\mathcal{K})=\max\limits_{\Delta \in \mathcal{K}} \rm{diam}(\Delta) $.  
	\end{definition}
	
	\begin{definition} [Size of Triangulation]
		Given a complex $\mathcal{K}$, \textit{the size of triangulation} is $\rm{card}(\mathcal{K})$. \textit{The size of smallest triangulation} is defined as $N(r)=\min\limits_{\rm{mesh}(\mathcal{K}) \leq r} \rm{card}(\mathcal{K})$.
	\end{definition}
	
	\begin{definition} [Polynomial Growth]
		A triangulable compact metric space $\mathbb{X}$ has polynomial growth on size of smallest triangulation if $N(r) \leq C_0/r^M, r>0$ where $C_0$ is a constant that is only related to $\mathbb{X}$.
	\end{definition}
	
	\begin{lemma} [Persistent Cycle Lemma]
		Let $X$ be a triangulable compact metric space and $f: X \rightarrow \mathbb{R}$ a tame Lipschitz function. The number of points in the persistence diagrams of $f$ whose persitence exceeds $\tau$ is at most $N(\tau/ \rm{Lip}(f))$. 
	\end{lemma}

	\begin{remark}
		The polynomial growth assumption is a natural assumption on the domain $\mathbb{X}$. For example, let $\mathbb{X}$ be a $d$-dimensional ball or cube we have $N(r)\leq C_0/ r^{d+\delta}$ for every $\delta >0$. 
	\end{remark}

\begin{lemma} [Bounded Persistence Moments] \label{lem_BSM}
		Let $\TotPers_k(f)=\sum_{p} \Pers(p)^k$ and assume the size of the smallest triangulation grows polynomially, i.e. $N(r)\leq C_0/r^M$. For $k>M$, there exists $C_X$ that only depends on $X$ such that $\TotPers_k(p)\leq C_X \rm{Lip}(f)^k$. 
	\end{lemma}

	\begin{lemma}[Lipschitz Constant of Persistence Moments] \label{lem_f3}
		Given two Lipschitz functions $f, f^\prime$ (they have the same Lipschitz constant, denoted as $Lip_x(f)$), there exist constants $c_1$ and $K$ such that 
		$$ |\TotPers_k(f)- \TotPers_k(f')| \leq c_1\|f-f'\|_{\infty} $$
		for every $k\geq K+1$. 
	\end{lemma}
	
	This bound could be derived from Total Persistence Stability Theorem in \cite{cohen2010lipschitz}. We include the proof here for completeness.
\begin{proof}
	By applying Lemma \ref{lem_BSM} with $k\geq d+3$  we have $\TotPers_{k-1}(f) \leq C_X\left(\rm{Lip}(f)\right)^{k-1}$.
	Using Stability Theorem in \cite{cohen2010lipschitz}, the persistence of points in diagrams of $f$ and $f'$ can be indexed into $m$ pairs where $m$ represents the maximum dots of diagrams of $f$ and $f'$. W.O.L.G when $f$ has more dots than $f'$, the  unpaired dots of $\dgm(f)$ are matched to zero persistence dot of $\dgm(f')$ (on the diagonal line). We have:\\
	\begin{equation*}
	\begin{aligned}
	&\TotPers_{k-1}(f)= \sum_{i=1}^{m} \Pers(p_i)^{k-1}   \\
	&\TotPers_{k-1}(f')= \sum_{i=1}^{m} \Pers(q_i) ^{k-1}  
	\end{aligned}
	\end{equation*}
	where $|\Pers(p_i)-\Pers(q_i)| \leq 2\|f-f'\|_{\infty}.$ Thus the persistence moment could be bounded as:
	\begin{equation}
    	\begin{aligned}
        	|\TotPers_k(f)-\TotPers_k(f')|&\leq \sum_{i=1}^{m} |\Pers(p_i)^k-\Pers(q_i)^k|\\
        	&\leq k \sum_{i=1}^m |\Pers(p_i)-\Pers(q_i)| \max\{\Pers(p_i),\Pers((q_i)\}^{k-1}\\
        	&\leq 2k \cdot \|f-f'\|_{\infty} \sum_{i=1}^{m} \max\{\Pers(p_i),\Pers(q_i)\}^{k-1}\\
        	&\leq 2k \cdot \|f-f'\|_{\infty}  (\TotPers_{k-1}(f)+\TotPers_{k-1}(f'))\\
        	&\leq 4k \cdot C_X\left(\rm{Lip}(f)\right)^{k-1} \|f-f'\|_{\infty}\\
        	&=c_1  \|f-f'\|_{\infty}
    	\end{aligned}
	\end{equation}
	Where $c_1 = 4k \cdot C_{X} (\rm{Lip}(f))^{k-1}$
\end{proof}	


\section{Proof of Lemma 1}
\label{sec:lemma1-proof}
\setcounter{lemma}{0}
\begin{lemma}
	Let's assume \textbf{A1} holds, we have
	We have:
	\begin{enumerate}
	    \item $\lambda_{topo}L_{topo}+\lambda_{reg}L_{reg}\le \lambda_{topo}{B} + \lambda_{reg}C_{ X}$. 
	    \item $\|\nabla_{W}(\lambda_{topo}L_{topo}+\lambda_{reg}L_{reg})\|_2\le 2\lambda_{reg} k C^2_X+ 2\lambda_{topo} k{B}$. 
	    \item $\|\nabla_{W}^2(\lambda_{topo}L_{topo}+\lambda_{reg}L_{reg})\|_2\le 2\lambda_{reg} (k(k-1)C_X+2kC_X)+ 2\lambda_{topo} k{B}$.
	\end{enumerate}

	\end{lemma}
\begin{proof}

	\textbf{(1)} The term $L_{reg}(W)=\sum_{p \in \dgm(f)} \Pers(p)^k \leq C_X(\rm{Lip}(f))^{k} = C_X$ (By Lemma \ref{lem_BSM} and the $1$-Lipschitz continuity of $f$ w.r.t. $x$ ). By \textbf{A1}, the term $L_{topo}$ is bounded by ${B}$ where ${B}$ is the number of selected dots of  $\truedgm$. Therefore, 
	$$\|\lambda_{topo}L_{topo}+\lambda_{reg}L_{reg}\| \leq \lambda_{reg} C_X + \lambda_{topo} {B}$$
	
	\textbf{(2)} We bound $\|\nabla _{W}L_{reg}(W)\|_2$ and $\|\nabla_{W}L_{topo}(W)\|_2$ respectively.
	The term 
	\begin{align*}
	    \|\nabla_{W}(L_{reg})\|_2 &= k\cdot\|\sum_{p \in \dgm(f)} \Pers(p)^{k-1}(\frac{\partial \death(p)}{\partial W} - \frac{\partial \birth(p)}{\partial W})\|_2 \\
	    & \leq 2k \TotPers_{k-1}(f)   \\
	    & \leq 2kC_X
	\end{align*}
	By \textbf{A1}, the term $\|\nabla_{W}L_{topo}\|_2$ is $2{B}$ bounded. Thus, the first order derivative is bounded: 
	$$\|\nabla_{W}(\lambda_{topo}L_{topo}+\lambda_{reg}L_{reg})\|_2\leq 2\lambda_{reg} k C_X+ 2\lambda_{topo} k{B}$$

	\textbf{(3)} We bound $\|\nabla ^2_{W}L_{reg}(W)\|_2$ and  $\|\nabla_{W}^2L_{topo}(W)\|_2$ respectively.
	\begin{equation*}
    	\begin{aligned}
        	\|\nabla^2 L_{reg}\|_2
        	& = k(k-1) \cdot \|\sum_{p \in \dgm( f)} \Pers(p)^{k-1}(\frac{\partial^2 \birth(p)}{\partial W\partial W^\top}-\frac{\partial^2 \death(p)}{\partial W\partial W^\top})\\
        	& + \sum_{p \in \dgm(f)} \Pers(p)^{k-2}(\frac{\partial \birth(p)}{\partial W}-\frac{\partial \death(p)}{\partial W})(\frac{\partial \birth(p)}{\partial W}-\frac{\partial \death(p)}{\partial W})^\top\|_2\\
        	& \leq 2k\TotPers_{k-1}(f) + 2k(k-1)\TotPers_{k-2}(f)  \\
        	& \leq 2kC_X + 2k(k-1)C_X 
    	\end{aligned}
	\end{equation*}
	Since $\|\nabla^2_{W}L_{topo}\|$  consists of quadratic functions, it  bounded by ${B}+4{B}$. In sum, the Hessian of our refined loss function can also be bounded: $$\|\nabla_{W}^2(\lambda_{topo}L_{topo}+\lambda_{reg}L_{reg})\|_2\leq 2\lambda_{reg} (k(k-1)C_X+2kC_X)+ 2\lambda_{topo} k{B}$$

\end{proof}

\end{document}

%% file: math_commands.tex
\usepackage{amsmath,amsfonts,bm}

\def\eqref#1{equation~\ref{#1}}

\def\1{\bm{1}}

\DeclareMathAlphabet{\mathsfit}{\encodingdefault}{\sfdefault}{m}{sl}
\SetMathAlphabet{\mathsfit}{bold}{\encodingdefault}{\sfdefault}{bx}{n}

\DeclareMathOperator*{\argmax}{arg\,max}

%% file: math_commands_cc.tex
\DeclareMathOperator{\dgm}{Dgm}
\DeclareMathOperator{\truedgm}{\overline{Dgm^\ast}}
\DeclareMathOperator{\orgtruedgm}{Dgm^\ast}
\DeclareMathOperator{\Pers}{Pers}
\DeclareMathOperator{\TotPers}{TotPers}

\DeclareMathOperator{\birth}{birth}
\DeclareMathOperator{\death}{death}
\DeclareMathOperator{\dist}{d}
\newcommand{\calD}{\mathcal{D}}
\DeclareMathOperator{\card}{card}

%% file: main.bbl
\begin{thebibliography}{10}

\bibitem{aukerman2020persistent}
Andrew Aukerman, Mathieu Carri{\`e}re, Chao Chen, Kevin Gardner, Ra{\'u}l
  Rabad{\'a}n, and Rami Vanguri.
\newblock Persistent homology based characterization of the breast cancer
  immune microenvironment: A feasibility study.
\newblock In {\em 36th International Symposium on Computational Geometry
  (SoCG)}, 2020.

\bibitem{carlsson2009topology}
Gunnar~E. Carlsson.
\newblock Topology and data.
\newblock {\em Bulletin of the American Mathematical Society}, 46:255--308,
  2009.

\bibitem{carriere2021optimizing}
Mathieu Carriere, Frederic Chazal, Marc Glisse, Yuichi Ike, Hariprasad Kannan,
  and Yuhei Umeda.
\newblock Optimizing persistent homology based functions.
\newblock In Marina Meila and Tong Zhang, editors, {\em Proceedings of the 38th
  International Conference on Machine Learning}, volume 139 of {\em Proceedings
  of Machine Learning Research}, pages 1294--1303. PMLR, 18--24 Jul 2021.

\bibitem{chen2019topological}
Chao Chen, Xiuyan Ni, Qinxun Bai, and Yusu Wang.
\newblock A topological regularizer for classifiers via persistent homology.
\newblock In {\em The 22nd International Conference on Artificial Intelligence
  and Statistics}, pages 2573--2582. PMLR, 2019.

\bibitem{clough2020topological}
James Clough, Nicholas Byrne, Ilkay Oksuz, Veronika~A Zimmer, Julia~A Schnabel,
  and Andrew King.
\newblock A topological loss function for deep-learning based image
  segmentation using persistent homology.
\newblock {\em IEEE Transactions on Pattern Analysis and Machine Intelligence},
  2020.

\bibitem{cohen2007stability}
David Cohen-Steiner, Herbert Edelsbrunner, and John Harer.
\newblock Stability of persistence diagrams.
\newblock {\em Discrete \& computational geometry}, 37(1):103--120, 2007.

\bibitem{cohen2010lipschitz}
David Cohen-Steiner, Herbert Edelsbrunner, John Harer, and Yuriy Mileyko.
\newblock Lipschitz functions have l p-stable persistence.
\newblock {\em Foundations of computational mathematics}, 10(2):127--139, 2010.

\bibitem{dey2022computational}
Tamal~Krishna Dey and Yusu Wang.
\newblock {\em Computational Topology for Data Analysis}.
\newblock Cambridge University Press, 2022.

\bibitem{edelsbrunner2010computational}
Herbert Edelsbrunner and John Harer.
\newblock {\em Computational Topology - an Introduction.}
\newblock American Mathematical Society, 2010.

\bibitem{edelsbrunner2000topological}
Herbert Edelsbrunner, David Letscher, and Afra Zomorodian.
\newblock Topological persistence and simplification.
\newblock In {\em Proceedings 41st annual symposium on foundations of computer
  science}, pages 454--463. IEEE, 2000.

\bibitem{gabrielsson2020topology}
Rickard~Br{\"u}el Gabrielsson, Bradley~J Nelson, Anjan Dwaraknath, and Primoz
  Skraba.
\newblock A topology layer for machine learning.
\newblock In {\em International Conference on Artificial Intelligence and
  Statistics}, pages 1553--1563. PMLR, 2020.

\bibitem{hatcher2000algebraic}
Allen Hatcher.
\newblock {\em {Algebraic topology}}.
\newblock Cambridge Univ. Press, Cambridge, 2000.

\bibitem{hofer2020topologically}
Christoph Hofer, Florian Graf, Marc Niethammer, and Roland Kwitt.
\newblock Topologically densified distributions.
\newblock In {\em International Conference on Machine Learning}, pages
  4304--4313. PMLR, 2020.

\bibitem{hofer2020graph}
Christoph Hofer, Florian Graf, Bastian Rieck, Marc Niethammer, and Roland
  Kwitt.
\newblock Graph filtration learning.
\newblock In {\em International Conference on Machine Learning}, pages
  4314--4323. PMLR, 2020.

\bibitem{hofer2019connectivity}
Christoph Hofer, Roland Kwitt, Marc Niethammer, and Mandar Dixit.
\newblock Connectivity-optimized representation learning via persistent
  homology.
\newblock In {\em International Conference on Machine Learning}, pages
  2751--2760. PMLR, 2019.

\bibitem{hofer2017deep}
Christoph Hofer, Roland Kwitt, Marc Niethammer, and Andreas Uhl.
\newblock Deep learning with topological signatures.
\newblock {\em Advances in neural information processing systems}, 30, 2017.

\bibitem{horn2021topological}
Max Horn, Edward De~Brouwer, Michael Moor, Yves Moreau, Bastian Rieck, and
  Karsten Borgwardt.
\newblock Topological graph neural networks.
\newblock {\em arXiv preprint arXiv:2102.07835}, 2021.

\bibitem{hu2019topology}
Xiaoling Hu, Fuxin Li, Dimitris Samaras, and Chao Chen.
\newblock Topology-preserving deep image segmentation.
\newblock {\em Advances in Neural Information Processing Systems}, 32, 2019.

\bibitem{hu2022trigger}
Xiaoling Hu, Xiao Lin, Michael Cogswell, Yi~Yao, Susmit Jha, and Chao Chen.
\newblock Trigger hunting with a topological prior for trojan detection.
\newblock In {\em International Conference on Learning Representations}, 2022.

\bibitem{hu2021topology}
Xiaoling Hu, Yusu Wang, Li~Fuxin, Dimitris Samaras, and Chao Chen.
\newblock Topology-aware segmentation using discrete morse theory.
\newblock In {\em International Conference on Learning Representations}, 2021.

\bibitem{jin2017escape}
Chi Jin, Rong Ge, Praneeth Netrapalli, Sham~M Kakade, and Michael~I Jordan.
\newblock How to escape saddle points efficiently.
\newblock In {\em International Conference on Machine Learning}, pages
  1724--1732. PMLR, 2017.

\bibitem{jin2021nonconvex}
Chi Jin, Praneeth Netrapalli, Rong Ge, Sham~M Kakade, and Michael~I Jordan.
\newblock On nonconvex optimization for machine learning: Gradients,
  stochasticity, and saddle points.
\newblock {\em Journal of the ACM (JACM)}, 68(2):1--29, 2021.

\bibitem{lawson2019persistent}
Peter Lawson, Andrew~B Sholl, J~Brown, Brittany~Terese Fasy, and Carola Wenk.
\newblock Persistent homology for the quantitative evaluation of architectural
  features in prostate cancer histology.
\newblock {\em Scientific reports}, 9(1):1--15, 2019.

\bibitem{munkers1984elements}
James~R. Munkres.
\newblock {\em {Elements of Algebraic Topology}}.
\newblock Addison Wesley Publishing Company, 1984.

\bibitem{nesterov2013introductory}
Yurii Nesterov.
\newblock {\em Introductory lectures on convex optimization: A basic course},
  volume~87.
\newblock Springer Science \& Business Media, 2013.

\bibitem{primoz2020wasserstein}
Primoz Skraba and Katharine Turner.
\newblock Wasserstein stability for persistence diagrams.
\newblock {\em arXiv:2006.16824v3}, 2020.

\bibitem{solomon2021fast}
Yitzchak Solomon, Alexander Wagner, and Paul Bendich.
\newblock A fast and robust method for global topological functional
  optimization.
\newblock In {\em International Conference on Artificial Intelligence and
  Statistics}, pages 109--117. PMLR, 2021.

\bibitem{wang2021topotxr}
Fan Wang, Saarthak Kapse, Steven Liu, Prateek Prasanna, and Chao Chen.
\newblock Topotxr: A topological biomarker for predicting treatment response in
  breast cancer.
\newblock In {\em International Conference on Information Processing in Medical
  Imaging}, pages 386--397. Springer, 2021.

\bibitem{wang2020topogan}
Fan Wang, Huidong Liu, Dimitris Samaras, and Chao Chen.
\newblock Topogan: A topology-aware generative adversarial network.
\newblock In {\em European Conference on Computer Vision}, pages 118--136.
  Springer, 2020.

\bibitem{wu2017optimal}
Pengxiang Wu, Chao Chen, Yusu Wang, Shaoting Zhang, Changhe Yuan, Zhen Qian,
  Dimitris Metaxas, and Leon Axel.
\newblock Optimal topological cycles and their application in cardiac
  trabeculae restoration.
\newblock In {\em International Conference on Information Processing in Medical
  Imaging}, pages 80--92. Springer, 2017.

\bibitem{wu2020topological}
Pengxiang Wu, Songzhu Zheng, Mayank Goswami, Dimitris Metaxas, and Chao Chen.
\newblock A topological filter for learning with label noise.
\newblock {\em Advances in neural information processing systems},
  33:21382--21393, 2020.

\bibitem{yan2021link}
Zuoyu Yan, Tengfei Ma, Liangcai Gao, Zhi Tang, and Chao Chen.
\newblock Link prediction with persistent homology: An interactive view.
\newblock In {\em International Conference on Machine Learning}, pages
  11659--11669. PMLR, 2021.

\bibitem{zhao2019learning}
Qi~Zhao and Yusu Wang.
\newblock Learning metrics for persistence-based summaries and applications for
  graph classification.
\newblock {\em Advances in Neural Information Processing Systems}, 32, 2019.

\bibitem{zhao2020persistence}
Qi~Zhao, Ze~Ye, Chao Chen, and Yusu Wang.
\newblock Persistence enhanced graph neural network.
\newblock In {\em International Conference on Artificial Intelligence and
  Statistics}, pages 2896--2906. PMLR, 2020.

\bibitem{zheng2021topological}
Songzhu Zheng, Yikai Zhang, Hubert Wagner, Mayank Goswami, and Chao Chen.
\newblock Topological detection of trojaned neural networks.
\newblock {\em Advances in Neural Information Processing Systems}, 34, 2021.

\bibitem{zomorodian2005computing}
Afra Zomorodian and Gunnar Carlsson.
\newblock Computing persistent homology.
\newblock {\em Discrete \& Computational Geometry}, 33(2):249--274, 2005.

\end{thebibliography}
